\documentclass[12pt]{article}

\usepackage{amssymb}
\usepackage{amsmath}
\usepackage{amscd}
\usepackage{latexsym}
\usepackage{graphicx}

\usepackage{caption}

\usepackage{cite}

\newcommand{\be}{\begin{equation}}
\newcommand{\ee}{\end{equation}}

\newcommand{\dlt}{\delta}
\newcommand{\prt}{\partial}

\newcommand{\bt}{\beta}

\newcommand{\ep}{\varepsilon}
\newcommand{\al}{\alpha}
\newcommand{\ra}{\rightarrow}

\begin{document}

\begin{center}

{\Large{\bf Discrete Versus Continuous Algorithms in Dynamics of Affective Decision Making
} \\ [5mm]

V.I. Yukalov$^{1,2}$ and E.P. Yukalova$^{3}$ }  \\ [3mm]

{\it
$^1$Bogolubov Laboratory of Theoretical Physics, \\
Joint Institute for Nuclear Research, Dubna 141980, Russia \\ [2mm]

$^2$Instituto de Fisica de S\~ao Carlos, Universidade de S\~ao Paulo, \\
CP 369, S\~ao Carlos 13560-970, S\~ao Paulo, Brazil \\ [2mm]

$^3$Laboratory of Information Technologies, \\
Joint Institute for Nuclear Research, Dubna 141980, Russia } \\ [3mm]

{\bf E-mails}: {\it yukalov@theor.jinr.ru}, ~~ {\it yukalova@theor.jinr.ru}

\end{center}

\vskip 2cm

\begin{abstract}
The dynamics of affective decision making is considered for an intelligent network 
composed of agents with different types of memory: long-term and short-term memory. 
The consideration is based on probabilistic affective decision theory, which takes into 
account the rational utility of alternatives as well as the emotional alternative 
attractiveness. The objective of this paper is the comparison of two multistep operational 
algorithms of the intelligent network: one based on discrete dynamics and the other
on continuous dynamics. By means of numerical analysis, it is shown that, depending on 
the network parameters, the characteristic probabilities for continuous and discrete 
operations can exhibit either close or drastically different behavior. Thus, depending 
on which algorithm is employed, either discrete or continuous, theoretical predictions 
can be rather different, which does not allow for a uniquely defined description of 
practical problems. This finding is important for understanding which of the algorithms 
is more appropriate for the correct analysis of decision-making tasks. A discussion is 
given, revealing that the discrete operation seems to be more realistic for describing 
intelligent networks as well as affective artificial intelligence.
\end{abstract}

\vskip 0.5cm
{\parindent=0pt
{\bf Keywords}: dynamic algorithms; affective decision making; dynamic decision 
making; affective artificial intelligence; probabilistic networks; intelligent 
networks }

\newpage

\section{Introduction}

Algorithms of modeling dynamic decision making are important for understanding and 
predicting the behavior of societies with regard to many principal problems that people
encounter in their life. As examples of such problems, it is possible to mention climate 
change, factory production, traffic control, firefighting, driving a car, military 
command, and so on. Research in dynamic decision making has focused on investigating the 
extent to which decision makers can use the obtained information and the acquisition of 
experience in making decisions. Dynamic decision making is a multiple, interdependent, 
real-time decision process, occurring in a changing environment. The latter can change 
independently or as a function of a sequence of actions by decision makers
\cite{Turkle_1,Brehmer_2,Beresford_3,Evertsz_4}. 

A society of decision makers forms a network, where separate agents play the role of 
network nodes. Decision making in networks has been studied in many papers that are
summarized in the recent reviews \cite{Perc_2013,Perc_2017,Capraro_2021,Jusup_2022}.
The role of moral preferences in following their personal and social norms has been
studied \cite{Capraro_2021}.   

Here, we consider dynamic decision making in a network of intelligent agents. The agents
make decisions in the frame of affective decision theory that is a probabilistic theory
where the agents choose alternatives taking account of both utility and emotions
\cite{Yukalov_5,Yukalov_6}. This theory can serve as a basis for creating affective 
artificial intelligence \cite{Yukalov_7}. The society of intelligent agents forms an 
intelligent network. Interactions between the agents occur through the exchange of 
information and through herding effect. 

Real-life situations are usually modeled by computer simulations, which is termed 
microworld modeling \cite{Turkle_1,Gonzalez_8}. The derivation of equations in dynamic
decision making can be achieved by assuming the time variation of an observable quantity in the 
presence of noise and then passing to the equations for the corresponding probabilities
\cite{Barendregt_9}. An important point in dynamic decision making is that living beings 
need to accumulate information adaptively in order to make sound decisions
\cite{Behrens_10,Ossmy_11}. This stresses the necessity of obtaining sufficient information 
for making optimal decisions. The received information accumulates in memory, which can
be of different types, say, long-term and short-term. Generally, the type of memory
depends on the environment and on the personality of decision makers. For example,
in quickly changing environments, animals use decision strategies that value recent 
observations more than older ones \cite{Yu_12,Brea_13,Urai_14}, although in gradually 
varying environments, they can have rather long-term memory. Human beings can have both 
types of memory, long-term and short-term memories \cite{Baddeley_15}.   

Decision making in a society of many agents includes several problems. One of them is 
associated with multi-agent reinforcement learning \cite{Albrecht_2023}. In the latter,
one considers a society of many agents in an environment shared by all members. The 
agents can accomplish actions leading to the change of the environmental state with 
a transition probability usually characterized by a Markov process. At each step of
the procedure, each agent receives an immediate reward, generally diminishing with time
due to time discounting. The aim of each agent is to find a behavioral policy, which
is a strategy that can guide the agent to take sequential actions that maximize 
the discounted cumulative reward.

The setup we consider has some analogies, although being quite different from 
multi-agent reinforcement learning. We consider a society where the environment 
for an agent consists of other society members. The state of the society is the set 
of probabilities of choosing alternatives by each member, with the probabilities taking 
account of the utility of alternatives as well as their attractiveness influencing the 
agents' emotions. The actions executed by the agents are the exchange of information 
on the choice of all other members. The aim of the agents is to find out whether 
stable distributions over the set of alternatives exist and, if so, what type of 
attractors they correspond to. The principal difference from multi-agent 
reinforcement learning is in two aspects: first, the aim is not a maximal reward, 
but a stable probability distribution over the given alternatives; and second, the 
influence of emotions is taken into account.  

Considering a sequence of multistep decision events, it is possible to accept two types 
of dynamics, based on either an algorithm with discrete time or with continuous time. 
The aim of the present paper is to compare these two kinds of algorithms in order to 
understand whether they are equivalent or not, and if they could lead to qualitatively
differing results. If it happens that the conclusions are principally different, it is 
necessary to decide which of the ways has to be used for the correct description of  
realistic situations. 

The layout of the paper is as follows. In order that the reader could better understand
the approach to affective decision making used in the present paper, it seems necessary
to recall the main points of this approach, which is presented in Section \ref{sec2} . In Section \ref{sec3}, the 
process of affective decision making in a society is formulated. In Section \ref{sec4}, the picture
is specified for a society composed of two groups of agents choosing between two 
alternatives in a multistep dynamics of decision making. One group of agents enjoys
long-term memory, while the other short-term memory. Section \ref{sec5} reformulates the 
dynamical process of multistep discrete decision making into a continuous process 
characterized by continuous time. In Section \ref{sec6}, a detailed numerical investigation is 
analyzed comparing the discrete and continuous algorithms of affective decision making. 
Section \ref{sec7} concludes.

\section{Affective Decision Making by Individuals}  \label{sec2} 

The usual approach to decision making is based on constructing a utility functional 
for each of the alternatives from the considered set \cite{Neumann_69,Savage_70}. In 
order to include the role of emotions, the expected utility is modified by adding the 
terms characterizing the influence of emotions \cite{Kurtz_71,Yaari_72,Reynaa_73,Woodford_74}. 
Thus, one tries to incorporate into utility at once both sides of decision making: 
rational reasoning, based on logical normative rules, and irrational unconscious 
emotions, such as joy, sadness, anger, fear, happiness, disgust, and surprise. The
alternative that corresponds to the largest expected utility is treated as optimal and
is certainly to be preferred.   

The approach we are using is principally different in several aspects: (i) This is 
a probabilistic theory, where the main characteristics are the probabilities of choosing
each of the given alternatives. (ii) The probability of a choice is the sum of a utility 
factor, describing the probability of a choice based on rational reasoning, and an 
attraction factor, characterizing the influence of emotions. (iii) The optimal, or more
correctly, a stochastically optimal alternative, is that which is associated with the 
largest probability.

The mathematically rigorous axiomatic formulation of the theory has been carried out in Refs. 
\cite{Yukalov_5,Yukalov_6,Yukalov_7}. The theory starts with the process of making 
decisions by separate individuals. Here, we state the main points of the approach in 
order that the reader could better understand the extension to decision making by 
a society, as is presented in this paper.

First of all, decision making is understood as a probabilistic process. Let us consider 
decision makers choosing between the alternatives from a set
\be
\label{01} 
 \mathbb{A} ~ = ~ \{ A_n: ~ n = 1,2, \ldots, N_A \} \;  .
\ee

The decision makers are considered as separate agents making decisions independently
from each other. Equivalently, it is possible to keep in mind a single decision maker
deciding on the given alternatives. The aim is to define the probability $p(A_n)$
of choosing an alternative $A_n$. This probability can be understood as either the
fraction of agents choosing this alternative or the frequency of choices of the 
alternative $A_n$ by a separate decision maker. Of course, the probability is 
normalized:
\be
\label{02}
 \sum_{n=1}^{N_A} p(A_n) ~ = ~ 1 \; , \qquad 
0 ~ \leq ~ p(A_n) ~ \leq ~ 1 \; .
\ee

The process of taking decisions consists of two sides. One evaluates the utility 
of alternatives as well as the attractiveness of alternatives that is influenced by 
emotions with respect to the choice of the alternatives. Therefore, the probability 
$p(A_n)$ of choosing an alternative $A_n$ is a behavioral probability consisting of 
two terms: a utility factor $f(A_n)$ and an attraction factor $q(A_n)$:
\be
\label{03}
p(A_n) ~ = ~ f(A_n) + q(A_n) \;   .
\ee

The utility factor $f(A_n)$ shows the rational probability of choosing an alternative 
$A_n$ being based on the rational evaluation of the alternative utility, with the 
normalization
\be
\label{04}
 \sum_{n=1}^{N_A} f(A_n) ~ = ~ 1 \; , \qquad 
0 ~ \leq ~ f(A_n) ~ \leq ~ 1 \; .
\ee

The attraction factor characterizes the influence of emotions in the process of choice
of the alternative $A_n$. Emotions can be positive or negative. For instance, the positive 
emotions are joy, happiness, pride, calm, serenity, love, gratitude, cheerfulness, 
euphoria, satisfaction (moral or physical), inspiration, amusement, pleasure, etc. Examples of
negative emotions are sadness, anger, fear, disgust, guilt, shame, anxiety, loneliness,
disappointment, etc. Taking into account Conditions (\ref{02})--(\ref{04}) 
implies 
\be 
\label{05}
  \sum_{n=1}^{N_A} q(A_n) ~ = ~ 0 \; , \qquad 
-1 ~ \leq ~ q(A_n) ~ \leq ~ 1 \;   .
\ee

To be more precise, the attraction factor varies in the interval
\be
\label{06}
 - f(A_n) ~ \leq ~ q(A_n) ~ \leq ~ 1 - f(A_n) \;   .
\ee

An alternative $A_{opt}$ is stochastically optimal if and only if it corresponds to the
maximal behavioral probability
\be
\label{07}
 p(A_{opt}) ~ = ~ \max_n p(A_n) \;  .
\ee

Let the alternatives be characterized by utilities (or value functionals) $U(A_n)$. The 
utility factor (rational probability) $f(A_n)$ can be derived from the minimization of 
the information functional 
\be
\label{08}
I[\; f(A_n) \; ]  ~ = ~ \sum_n f(A_n) \; \ln \frac{f(A_n)}{f_0(A_n)}
+ \al \left[ \; 1 - \sum_n f(A_n) \; \right] +
\bt \left[ \; U - \sum_n f(A_n) U(A_n) \; \right]  \; ,
\ee
where $f_0(A_n)$ is a prior distribution defined by the Luce rule \cite{Luce_75,Luce_76}, 
which gives
\be
\label{09}
 f(A_n)  ~ = ~ 
\frac{f_0(A_n) e^{\bt U(A_n)} }{\sum_n f_0(A_n) e^{\bt U(A_n)} } \; .
\ee

The parameter $\beta$ is a belief parameter characterizing the level of certainty of 
a decision maker in the fairness of the decision task and in the subject confidence 
with respect to their understanding of the overall rules and conditions of the
decision problem \cite{Yukalov_5,Yukalov_6,Yukalov_7}. Here, we keep in mind rational 
beliefs representing reasonable, objective, flexible, and constructive conclusions or 
inferences about reality \cite{Brandt_77,Swinburne_78}.

The attraction factor is a random quantity that is different for different decision 
makers and even for the same decision maker at different times. The average values
for positive or negative emotions of the attraction factor can be estimated by 
non-informative priors as $\pm 0.25$, respectively \cite{Yukalov_6,Yukalov_7}. The 
description of decision making by independent agents in the frame of probabilistic
affective decision making has been studied and expounded in detail in Refs. 
\cite{Yukalov_5,Yukalov_6,Yukalov_7}. The aim of the present paper is to consider the
extension of the theory from single-step affective decision making of a single agent
to multistep dynamic affective decision making by a society of many decision makers.  

Utility factors are objective quantities that can be calculated provided the utility
of alternatives $U(A_n)$ are defined. Generally, $U(A_n)$ can be an expected utility,  
a value functional, or any other functional measuring the rational utility of alternatives.
For example, in the case of multi-criteria decision making, this can be an objective 
function defined by one of the known multi-criteria evaluation methods 
\cite{Steuer_79,Triantaphyllou_80,Koksalan_81,Basilio_82}. For the purpose of the present 
paper, we do not need to plunge into numerous methods of evaluating the utility of 
alternatives. We assume that the utility factor is defined in one of the ways. Our basic 
goal is the investigation of the role of emotions.    

In what follows, we assume that the utility factors, evaluated at the initial moment of 
time, do not change, since their values have been objectively defined. On the contrary, 
The attraction factors depend on emotions that change in the process of decision 
making due to the exchange of information between the society members and because the 
behavior of decision makers is influenced by the actions of other members of a society.

\section{Discrete Dynamics in Affective Decision Making} \label{sec3}

The approach to affective decision making, considered in the present paper, is based on 
the probabilistic theory \cite{Yukalov_5,Yukalov_6,Yukalov_7} characterized by probabilities 
of choosing an alternative among the set of given alternatives, taking account of 
utility as well as emotions. In studying dynamic equations, one has to define initial 
conditions, that is, the utility factors and attraction factors at time $t=0$. At the initial 
time, the decisions are taken by agents independently, since they have no time for 
exchanging information and observing the behavior of their neighbors. Thus, the initial 
behavioral probabilities define the required initial conditions for the following dynamics. 
 
A society, or a network, is considered to consist of many agents. For each member of a 
society, the other members play the role of surrounding. The agents of a society interact 
with each other through the exchange of information and by imitating the actions of others. 
The probability dynamics is due to these features \cite{Yukalov_16,Yukalov_17,Yukalov_18}.
 
Let us consider $N_A$ alternatives between which one needs to make a choice. The 
alternatives are enumerated by the index $n = 1,2,\ldots,N_A$. A society of $N_{tot}$ 
agents is making a choice among the available alternatives. The overall society is 
structured into $N$ groups enumerated by the index $j = 1,2,\ldots,N$. Each group differs 
from other groups by its specific features, such as its type of memory and the inclination 
to replicate the actions of others, which is termed {\it herding}. The herding effect 
is well known and has been studied in voluminous literature \cite{Martin_19,Sherif_20,Smelser_21,
Merton_22,Turner_23,Hatfield_24,Brunnermeier_25,Sornette_26,Yukalov_27}. 

The number of agents in a group $j$ is $N_j$ so that the summation over all groups gives
the total number of agents,
\be
\label{1}
 \sum_{j=1}^N N_j ~ = ~ N_{tot} \;  .
\ee

The number of agents in a group $j$ choosing an alternative $A_n$ at time $t$ is 
$N_j(A_n,t)$. Since each member of a group $j$ chooses one alternative, then
\be
\label{2}
  \sum_{n=1}^{N_A} N_j(A_n,t) ~ = ~ N_j \;  .
\ee

The probability that a member of a group $j$ chooses an alternative $A_n$ at time $t$ is
\be
\label{3}
  p_j(A_n,t) ~ \equiv ~ \frac{N_j(A_n,t)}{N_j} \;  ,
\ee
which satisfies the normalization condition
\be
\label{4}
 \sum_{n=1}^{N_A} p_j(A_n,t) ~ = ~ 1 \; , \qquad 
0 ~ \leq ~ p_j(A_n,t) ~ \leq ~ 1 \;  .
\ee
    
Probability (\ref{3}) is a functional of the utility factor $f_j(A_n,t)$ and the 
attraction factor $q_j(A_n,t)$. The utility factor characterizes the utility of an 
alternative $A_n$ at time $t$ and obeys the normalization condition
\be
\label{5}
  \sum_{n=1}^{N_A} f_j(A_n,t) ~ = ~ 1 \; , \qquad 
0 ~ \leq ~ f_j(A_n,t) ~ \leq ~ 1 \;   .
\ee

The attraction factor quantifies the influence of emotions when selecting an alternative
$A_n$ at time $t$ and satisfies the normalization condition
\be
\label{6}
 \sum_{n=1}^{N_A} q_j(A_n,t) ~ = ~ 0 \; , \qquad 
-1 ~ \leq ~ q_j(A_n,t) ~ \leq ~ 1 \;    .
\ee

At the initial moment of time $t=0$, the functional dependence of the probability on the 
utility and attraction factors has the form
\be
\label{7}
 p_j(A_n,0) ~ = ~ f_j(A_n,0) + q_j(A_n,0) \; ,
\ee
where the initial utility factor and attraction factor can be calculated following the 
rules explained in detail in earlier works \cite{Yukalov_5,Yukalov_6,Yukalov_7,Yukalov_27,
Yukalov_28,Yukalov_29}.   

The tendency of agents of a group $j$ to replicate the actions of the members of other
groups is described by the {\it herding parameters} $\varepsilon_j$, which lay in the 
interval
\be
\label{8}
0 ~ \leq ~ \ep_j ~ \leq ~ 1 \qquad ( j = 1,2,\ldots,N) \; .
\ee  

The other meaning of these parameters is the level of tendency for acting as others, which
in the present setup models the agents' cooperation.

Generally, the value $\varepsilon_j$ can vary in time. However, this variation is usually
very slow so that the herding parameters can be treated as constants characterizing the
members of the related groups.

The time evolution, consisting of a number of subsequent decisions at discrete moments of 
time $t/\tau = 1,2,\ldots$, is given by the dynamic equation
$$
p_j(A_n,t+\tau) ~ = ~ (1 - \ep_j) \; [\; f_j(A_n,t) + q_j(A_n,t) \; ] \; + 
$$
\be
\label{9}
 +\;
\frac{\ep_j}{N-1} \sum_{i(\neq j)}^N [\; f_i(A_n,t) + q_i(A_n,t) \; ] \; ,
\ee
where $\tau$ is a delay time required for taking a decision by an agent. It is possible
to measure time in units of $\tau$ keeping in mind the dimensionless time $t=1,2,\ldots$. 
The time dependence of the utility factor can be prescribed by a discount function 
\cite{Yukalov_7,Read_30,Frederick_31}, and the temporal dependence of the attraction 
factor for an agent of a group $j$,
\be
\label{10}
 q_j(A_n,t) ~ = ~ q_j(A_n,0) \exp\{ - M_j(t) \; \},  
\ee
is defined by the amount of information received from other society members and kept in 
the memory $M_j(t)$ by time $t$. The derivation of Relation (\ref{10}) can be achieved by 
resorting to the theory of quantum measurements \cite{Yukalov_53} or by accepting
the empirical fact \cite{Kuhberger_54,Charness_55,Cooper_56,Blinger_57,Sutter_58,Tsiporkova_59,
Charness_60,Charness_61,Chen_62,Liu_63,Charness_64,Sung_65,Schultze_66,Xu_67,Tapia_68}
that the increase in information kept in the memory decreases the role of emotions so
that $\delta q_j = -q_j \delta M_j$.        

At the beginning, when $t < 1$, there is no yet any memory with respect to the choice
between the present alternatives so that 
\be
\label{11}
M_j(t)  ~ = ~ 0 \qquad ( t < 1) \;   ,
\ee
and one returns to the initial condition (\ref{7}). For the time $t\geqslant 1$, the 
memory is written as
\be
\label{12}
 M_j(t)  ~ = ~ 
\Theta(t-1) \sum_{t'=1}^t \; \sum_{i=1}^N J_{ij}(t,t') \mu_{ji}(t') \;  ,
\ee
where $J_{ij}(t,t')$ is the interaction transfer function describing the interaction 
between the agents $i$ and $j$ during the time from $t'$ to $t$, $\mu_{ji}$ is 
the information gain received by the agent $j$ from the agent $i$, and the unit-step 
function is used
\begin{eqnarray}
\nonumber
\Theta(t-1) = \left\{ \begin{array}{ll}
0 \; , ~ & ~ t < 1 \\
1 \; , ~ & ~ t \geq 1 \end{array} 
\; .
\right.
\end{eqnarray}

In contemporary societies, the interaction between agents is of long-range type, 
since the society members are able to interact by exchanging information through 
numerous sources not depending on the distance, e.g. through phone, Skype, Whatsapp, 
and a number of other messengers. The long-range interactions are characterized by 
the expression
\be
\label{13}
 J_{ij}(t,t') ~ = ~ \frac{J(t,t')}{N-1} \; .
\ee

On the contrary, in the case of short-range interactions, $J_{ij}(t,t')$ essentially 
depends on the fixed location of agents. However, the members of modern societies are 
not fixed forever to precise prescribed locations. This concerns not only human societies, 
but animal groups as well. Therefore, the long-range interaction (\ref{13}) looks to be
the most realistic case. 

The information gain can be taken in the Kullback--Leibler \cite{Kullback_32,Kullback_33}  
form
\be
\label{14}
 \mu_{ji}(t,t') ~ = ~  \sum_{n=1}^{N_A} p_j(A_n,t) \; 
\ln \frac{p_j(A_n,t)}{p_i(A_n,t)} \;  .
\ee

Thus, the memory function (\ref{12}) reads as 
\be
\label{15}
 M_j(t) ~ = ~ \Theta(t-1) \sum_{t'=1}^t \frac{J(t,t')}{N-1} \;
\sum_{i=1}^N \mu_{ji}(t') \; .
\ee
  
From the point of view of duration, there exist two types of memory: long-term and 
short-term memory \cite{Baddeley_15,James_34,Fitts_35,Cowan_36,Camina_37}. Long-term memory 
allows us to store information for long periods of time, including information that can be 
retrieved. This implies weak dependence of the interaction transfer on time,
\be
\label{16}
  J(t,t') ~ = ~ J \qquad (long - term) \;  ,
\ee
which defines the long-term memory
\be
\label{17}
  M_j(t) ~ = ~ \Theta(t-1) \frac{J}{N-1} \sum_{t'=1}^t \;
\sum_{i=1}^N \mu_{ji}(t') \; .
\ee

Short-term memory is the capacity to store a small amount of information in the mind and 
keep it readily available for a short period of time. Then, the interaction transfer is 
modeled by the function
\be
\label{18}
 J(t,t') ~ = ~ J \dlt_{tt'} \qquad (short - term) \; ,
\ee
so that the short-term memory takes the form
\be
\label{19}
 M_j(t) ~ = ~ \Theta(t-1) \frac{J}{N-1} \sum_{i=1}^N \mu_{ji}(t) \;  .
\ee

\section{Two Groups with Binary Choice} \label{sec4} 

For concreteness, let us study the case where the choice is between two alternatives,
$A_1$ and $A_2$. Then, it is convenient to simplify the notation by setting the 
probabilities
\be
\label{20}
 p_j(A_1,t) ~ \equiv ~  p_j(t) \; , \qquad p_j(A_2,t) ~ = ~ 1 - p_j(t) \; ,
\ee
the utility factors
\be
\label{21}
 f_j(A_1,t) ~ \equiv ~  f_j(t) \; , \qquad f_j(A_2,t) ~ = ~ 1 - f_j(t) \;  ,
\ee
and the attraction factors
\be
\label{22}
q_j(A_1,t) ~ \equiv ~  q_j(t) \; , \qquad q_j(A_2,t) ~ = ~ - q_j(t) \;   ,
\ee
where the normalization conditions (\ref{4})--(\ref{6}) are taken into 
account. 
 
Let the society consist of two groups, one whose members possess long-term memory
and the other group consisting of the members with short-term memory. In the following 
numerical modeling, we set $J = 1$. Now, the long-term memory reads as
\be
\label{23}
M_1(t) ~ = ~ \Theta(t-1) \sum_{t'=1}^t \mu_{12}(t') \qquad (long - term) \;  ,
\ee
while the short-term memory becomes
\be
\label{24}
M_2(t) ~ = ~ \Theta(t-1)  \mu_{21}(t') \qquad (short - term) \;   .
\ee

The information gain (\ref{14}) takes the form
\be
\label{25}
 \mu_{ij}(t)  ~ = ~ p_i(t) \; \ln \; \frac{p_i(t)}{p_j(t)} +
[\; 1 - p_i(t) \; ] \; \ln \; \frac{1-p_i(t)}{1-p_j(t)} \; .
\ee

For brevity, let us use the notations
\be
\label{26}
f_j(0) ~ \equiv ~ f_j \; , \qquad q_j(0) ~ \equiv ~ q_j \;  .
\ee

Also, we assume that the process of making decisions concerns the alternatives with given
utilities so that
\be
\label{27}
 f_j(t) ~ = ~ f_j  ~ = ~ const \; ,
\ee
although emotions can vary due to the exchange of information between the agents.  

Thus, we come to the equations of dynamic decision making
$$
p_1(t+1) ~ = ~ (1-\ep_1) [\; f_1 + q_1(t) \; ] + \ep_1 [\; f_2 + q_2(t) \; ] \; ,
$$
\be
\label{28}
 p_2(t+1) ~ = ~ (1-\ep_2) [\; f_2 + q_2(t) \; ] + \ep_2 [\; f_1 + q_1(t) \; ] \;  ,
\ee
with the initial conditions
\be
\label{29}
 p_1(0) ~ = ~ f_1 + q_1 \; , \qquad p_2(0) ~ = ~ f_2 + q_2 \;  .
\ee

The attraction factors have the form
\be
\label{30}
q_1(t) ~ = ~ q_1 \exp\{-M_1(t) \} \; , \qquad
 q_2(t) ~ = ~ q_2 \exp\{-M_2(t) \} \;  ,
\ee
with the long- and short-term memories (\ref{23}) and (\ref{24}).

\section{Continuous Dynamics of Affective Decision Making} \label{sec5} 

Repeated multistep decision making is a discrete process, as is described above. 
However, if the time of taking a decision is much shorter than the whole multistep 
process, $\tau/t \ll 1$, then it looks admissible to pass from the equations with
discrete time to continuous time by expanding the probabilities in powers of $\tau/t$,
\be
\label{31}
 p_j(A_n,t+\tau) ~ \simeq ~ 
p_j(A_n,t) + \frac{\prt p_j(A_n,t)}{\prt t} \; \tau \;  .
\ee

Measuring time again in units of $\tau$ gives
\be
\label{32}
  p_j(A_n,t+1) ~ \simeq ~ 
p_j(A_n,t) + \frac{\prt p_j(A_n,t)}{\prt t} \; .
\ee

Using this, Equation (\ref{9}) transforms into
$$
 \frac{\prt p_j(A_n,t)}{\prt t}  ~ = ~  
(1 - \ep_j) [\; f_j(A_n,t) + q_j(A_n,t) \; ] \; +
$$
\be
\label{33}
+\;
 \frac{\ep_j}{N-1} 
\sum_{i(\neq j)}^N [\; f_i(A_n,t) + q_i(A_n,t) \; ] - p_j(A_n,t) \; .
\ee

For the binary case of the previous section, we obtain 
$$
\frac{d p_1(t)}{d t}  ~ = ~ (1 - \ep_1) [\; f_1 + q_1(t) \; ] +
\ep_1 [\; f_2 + q_2(t) \; ] - p_1(t) \; ,
$$
\be
\label{34}
\frac{d p_2(t)}{d t}  ~ = ~ (1 - \ep_2) [\; f_2 + q_2(t) \; ] +
\ep_2 [\; f_1 + q_1(t) \; ] - p_2(t) \;  .
\ee

For small $\tau$, it is possible to use the relation
$$
\sum_{t'=\tau}^t \mu_{12}(t') ~ \simeq ~ 
\int_\tau^t \mu_{12}(t') \; dt' \qquad (\tau\ra 0, ~ t \geq \tau ) \; ,
$$
which yields the long-term memory
\be
\label{35}
 M_1(t) ~ = ~ \int_0^t \mu_{12}(t') \; dt' \; .
\ee

Employing the approximate equality
$$
\Theta(t-\tau) ~ \simeq ~ \tanh\left( \frac{t}{\tau} \right) \qquad
(\tau \ra 0 \; , ~ t \geq 0 ) \;   ,
$$
the short-term memory can be represented as
\be
\label{36}
  M_2(t) ~ = ~ \tanh\left( \frac{t}{\tau} \right) \; \mu_{21}(t) \; .
\ee

In numerical calculations, $\tau$ is taken as a step of the used numerical scheme.

\section{Comparison of Discrete Versus Continuous Algorithms}  \label{sec6} 

Formally, it looks that the fixed points, if they exist, of the discrete (\ref{28}) 
and continuous (\ref{34}) dynamical systems are the same, being given by the equations
$$
 p_1^* ~ = ~  
(1 - \ep_1) ( f_1 + q_1^* ) + \ep_1 ( f_2 + q_2^* ) \; , 
$$
\be
\label{37}
 p_2^* ~ = ~  
(1 - \ep_2) ( f_2 + q_2^* ) + \ep_2 ( f_1 + q_1^* ) \; ,
\ee
where $q^*_j$ is the limit of $q_j(t)$ as time goes to infinity. However, strictly 
speaking, the discrete and continuous limits can be different, since the related 
expressions for the memory functions in the discrete and continuous cases are different.
Also, the considered equations are not autonomous and contain time delay. In addition, 
even if the fixed points would be the same, the stability conditions of discrete, 
continuous, and delay equations, generally, are different 
\cite{Gershenfeld_38,Matsumoto_39,Yukalov_40}. Thus, numerical investigations are 
necessary.

We have compared the solutions to discrete-time Equation (\ref{28}) and continuous-time 
Equation (\ref{34}), for the same sets of parameters and initial conditions. The society 
is composed of two groups, one whose members enjoy long-term memory and the other group, 
consists of members with short-term memory. Solutions for discrete equations are marked 
as $p_j^{dis}(t)$ and for continuous equations as $p_j^{con}(t)$. In all figures,
time is dimensionless, being measured in units of $\tau$. The results are discussed below.

\begin{figure}[!ht]
\centering 
\includegraphics[width=8cm]{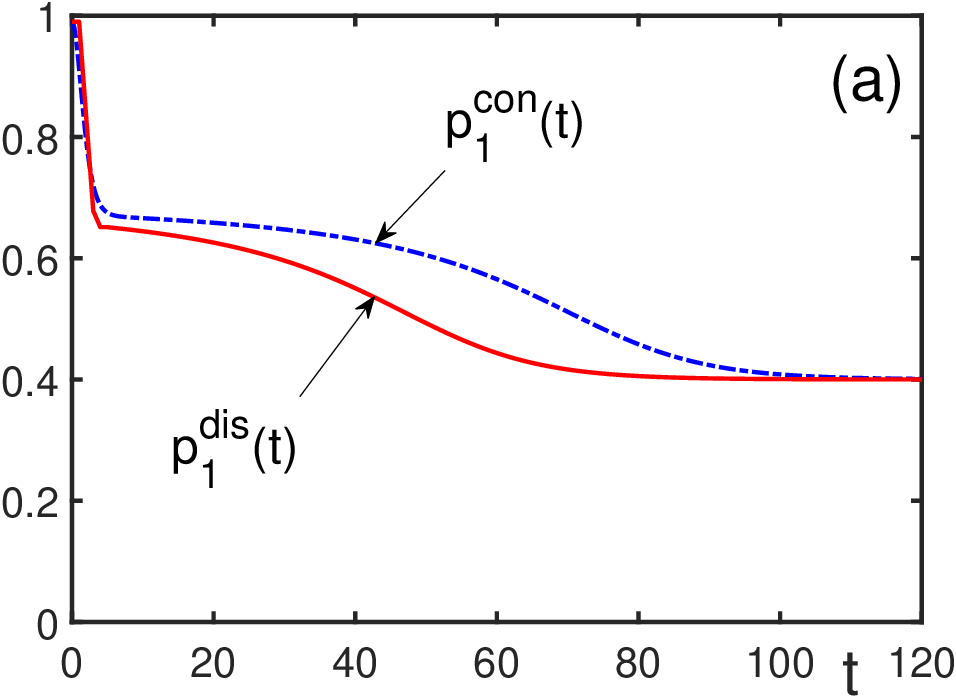} \hspace{0.5cm}
\includegraphics[width=8cm]{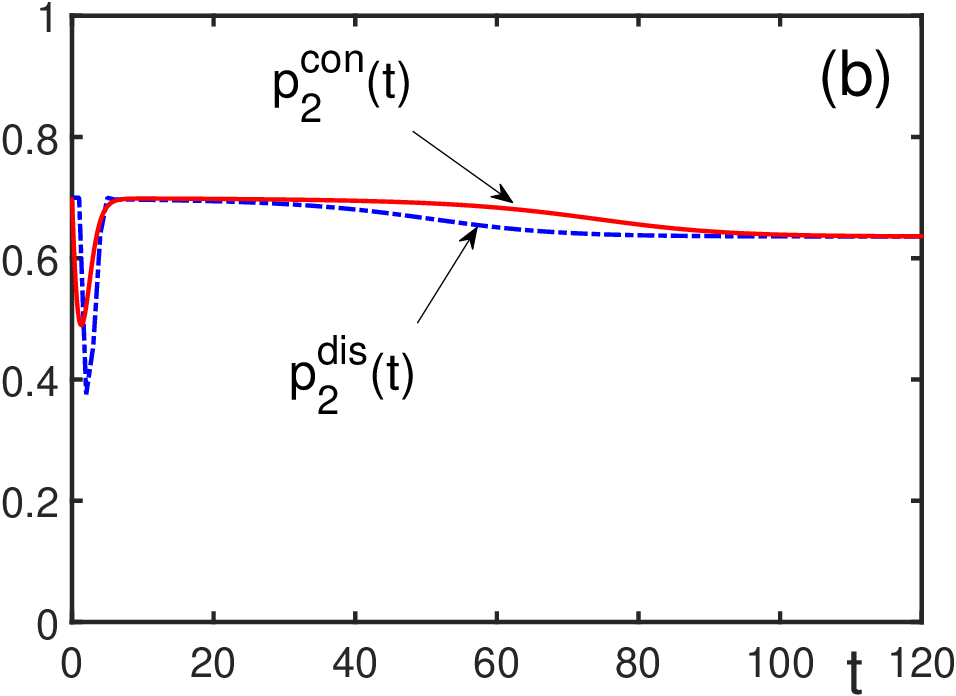}
\caption{\small
Solutions to discrete Equation (\ref{28}) and to continuous Equation (\ref{34}) 
for the initial conditions $f_1=0.4$, $f_2=0.1$, $q_1=0.59$, and $q_2=0.6$, in the 
absence of herding effect, when $\ep_1=\ep_2=0$:
(\textbf{a}) 
Discrete solution $p_1^{dis}(t)$ (solid line) and continuous solution $p_1^{con}(t)$ 
(dashed-dotted line). Both solutions tend to the same fixed point $p_1^*=0.4$; 
(\textbf{b}) 
Discrete solution $p_2^{dis}(t)$ (solid line) and continuous solution $p_2^{con}(t)$ 
(dashed-dotted line). Both solutions tend to the same fixed point $p_2^*=0.636$, which 
is a stable node.
}
\label{fig:Fig.1}
\end{figure}

Figure \ref{fig:Fig.1} presents the case where the fractions (probabilities) $p_j^{con}(t)$ 
and $p_j^{dis}(t)$ starting from the same values smoothly tend to the same fixed points, 
being only slightly different at intermediate times.    

\begin{figure}[!ht]
\centering
\includegraphics[width=8cm]{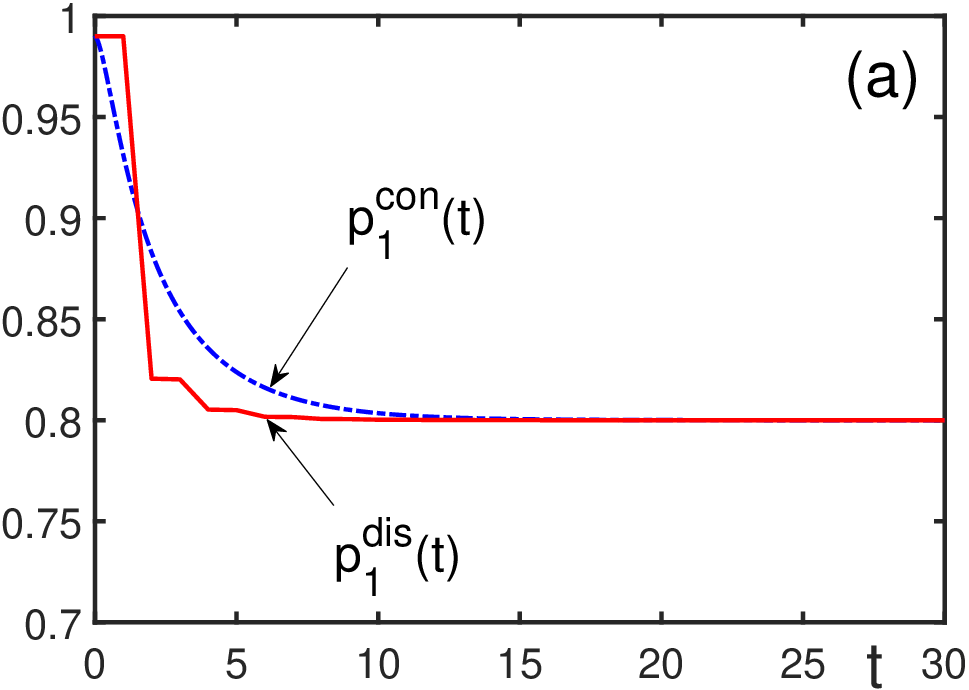} \hspace{0.5cm}
\includegraphics[width=8cm]{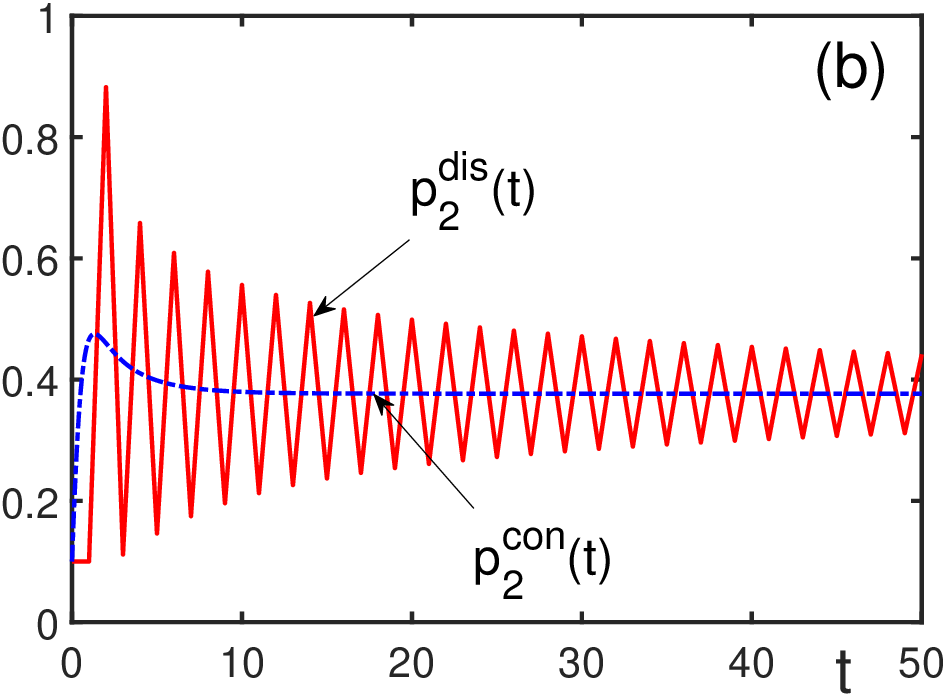}
\caption{\small
Solutions to discrete Equation (\ref{28}) and to continuous Equation 
(\ref{34}) for the initial conditions $f_1=0.8$, $f_2=0.9$, $q_1=0.19$, and 
$q_2=-0.8$, when there is no herding effect, hence $\ep_1=\ep_2=0$:
(\textbf{a}) Discrete solution $p_1^{dis}(t)$ (solid line) and continuous solution 
$p_1^{con}(t)$ (dashed-dotted line). Both solutions tend to the same fixed point 
$p_1^*=0.8$;
(\textbf{b}) Discrete solution $p_2^{dis}(t)$ (solid line) and continuous solution 
$p_2^{con}(t)$ (dashed-dotted line). Probability $p_2^{con}(t)$ tends monotonically, 
while $p_2^{dis}(t)$ tends with oscillations to the same fixed point $p_2^*=0.377$. 
Discrete and continuous solutions tend to the same fixed point, but for the agents 
with long-term memory the fixed point is a stable node, however for the agents with 
short-term memory, the continuous solution tends to a stable node, while for the 
discrete solution, to a stable focus.
}
\label{fig:Fig.2}
\end{figure}

Figure \ref{fig:Fig.2} shows the situation when the probabilities of choosing an 
alternative by agents with long-term memory smoothly tend to the same fixed point, 
but the probabilities for agents with short-term memory, although tending to the 
same fixed point, tend in a rather different way. The continuous solution tends 
smoothly, while the discrete solution, through oscillations.

\begin{figure}[!ht]
\centering\includegraphics[width=8cm]{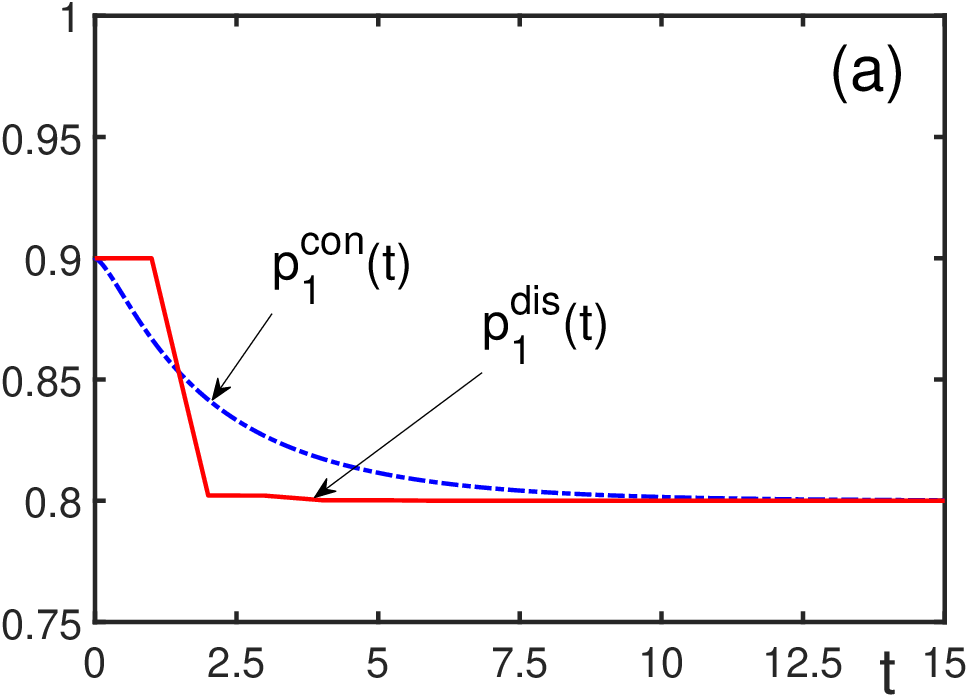} \hspace{5mm}
\includegraphics[width=8cm]{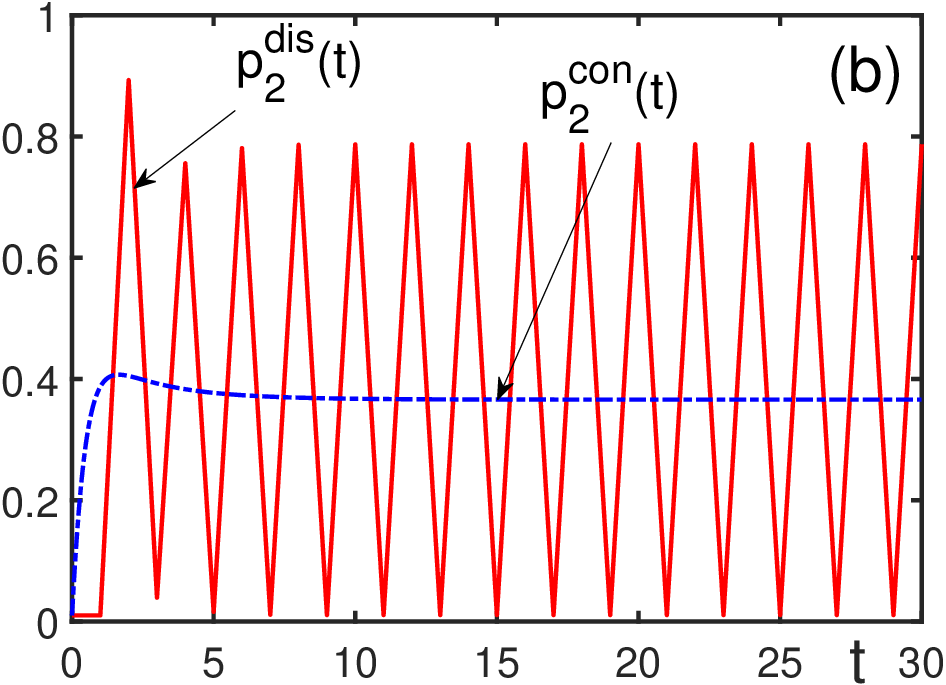}
\caption{\small
Solutions to discrete Equation (\ref{28}) and to continuous Equation 
(\ref{34}) for the initial conditions $f_1=0.8$, $f_2=1$, $q_1=0.1$, and $q_2=-0.99$, 
in the absence of herding effect, when $\ep_1=\ep_2=0$:
(\textbf{a}) Discrete solution $p_1^{dis}(t)$ (solid line) and continuous solution 
$p_1^{con}(t)$ (dashed-dotted line). Solutions $p_1^{con}(t)$ and $p_1^{dis}(t)$ 
tend to the same fixed point $p_1^*=0.8$;
(\textbf{b}) Discrete solution $p_2^{dis}(t)$ (solid line) and continuous solution 
$p_2^{con}(t)$ (dashed-dotted line). Solution $p_2^{con}(t)$ tends to the fixed point 
$p_2^*=0.366$, whereas $p_2^{dis}(t)$ oscillates around $p_2^*$ with the constant 
amplitude. For the agents with long-term memory, both probabilities, discrete and 
continuous, tend to the same stable node, but for the agents with short-term memory, 
the fixed point for discrete probability is a stable limit cycle, while the continuous 
probability tends to a stable node. 
}
\label{fig:Fig.3}
\end{figure}

Figure \ref{fig:Fig.3} demonstrates that the fixed points of discrete and continuous 
solutions can be of different nature. Thus, for the group of agents with long-term 
memory, the discrete and continuous solutions tend to the same stable node. However, 
for the agents with short-term memory, it is a stable node for the continuous solution, 
but a center for the discrete solution.

\begin{figure}[!ht]
\centering\includegraphics[width=8cm]{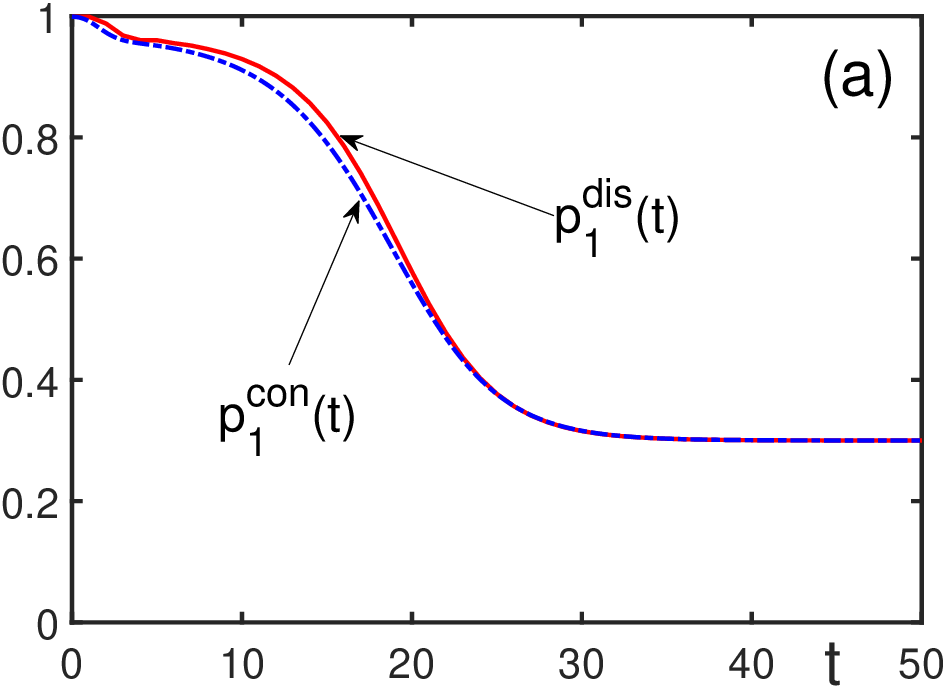} \hspace{5mm}
\includegraphics[width=8cm]{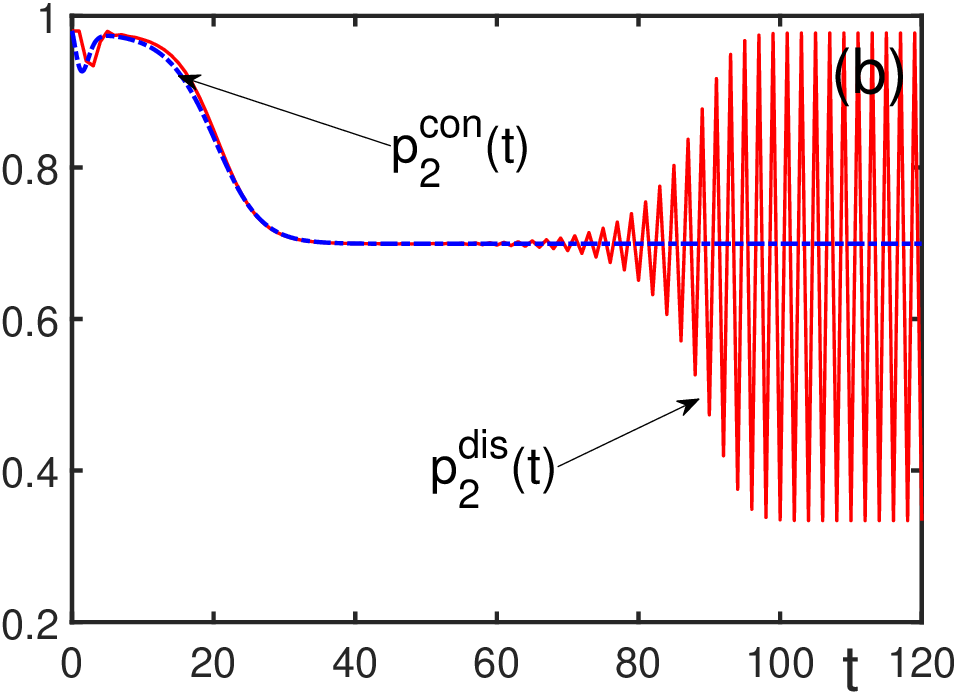}
\caption{\small
Solutions to discrete Equation (\ref{28}) and to continuous Equation 
(\ref{34}) for the initial conditions $f_1=0.3$, $f_2=0$, $q_1=0.699$, and 
$q_2=0.98$, without the herding effect, when $\ep_1=\ep_2=0$:
(\textbf{a}) Discrete solution $p_1^{dis}(t)$ (solid line) and continuous solution 
$p_1^{con}(t)$ (dashed-dotted line). Solutions $p_1^{con}(t)$ and $p_1^{dis}(t)$ 
tend to the same fixed point $p_1^*=0.3$;
(\textbf{b}) Discrete solution $p_2^{dis}(t)$ (solid line) and continuous solution 
$p_2^{con}(t)$ (dashed-dotted line). Solution $p_2^{con}(t)$ tends to $p_2^*=0.699$, 
whereas $p_2^{dis}(t)$ oscillates around $p_2^*$, starting at a finite time and 
continues oscillating for $t\ra\infty$ with a constant amplitude. The fixed points 
of agents with long-term memory coincide for discrete and continuous solutions, 
both being stable nodes, while for agents with short-term memory, the continuous 
solution tends to a stable node, whereas the discrete one oscillates.
}
\label{fig:Fig.4}
\end{figure}

Figure \ref{fig:Fig.4} shows that the fixed points of agents with long-term memory 
can coincide for discrete and continuous solutions, both being stable nodes, while 
for agents with short-term memory, the continuous solution tends to a stable node, 
whereas the discrete solution at the beginning almost coincides with the continuous 
one, but starts oscillating from a finite time and after this continues oscillating 
for all times.   

\begin{figure}[!ht]
\centering
\includegraphics[width=8cm]{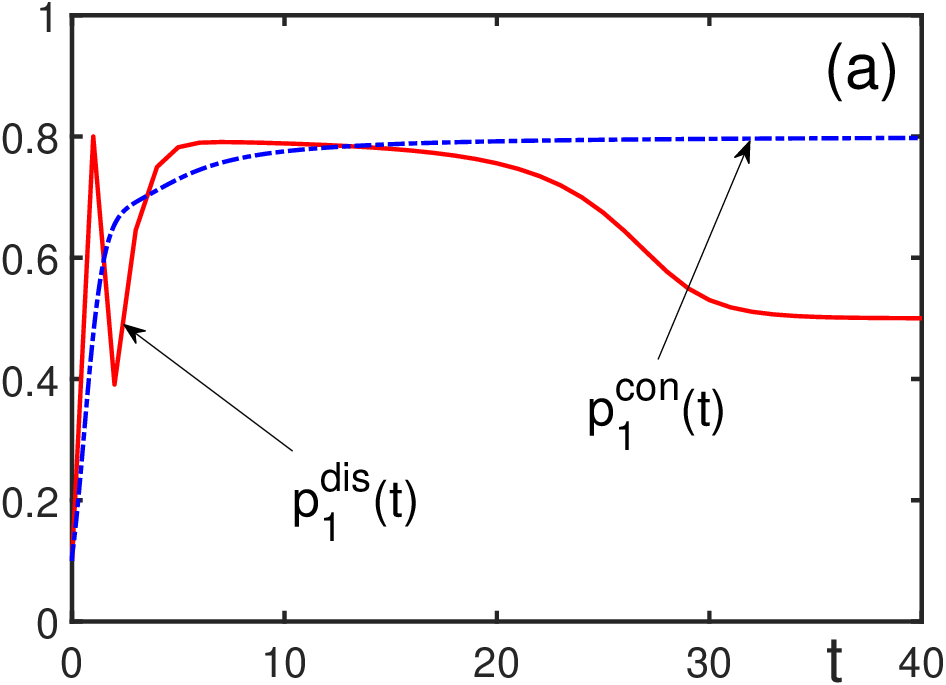} \hspace{5mm}
\includegraphics[width=8cm]{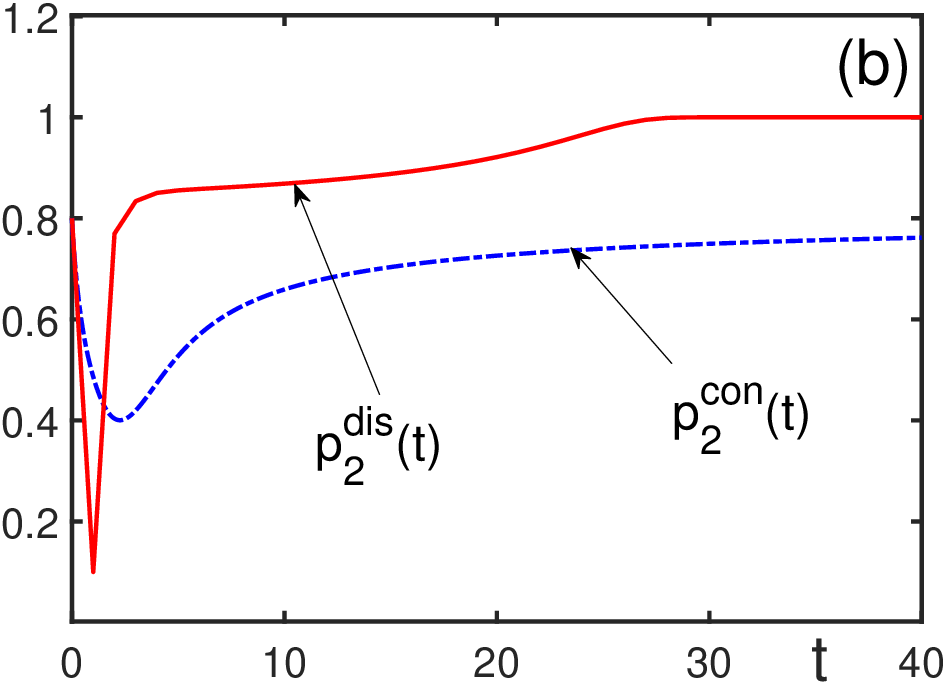}\\
\vskip 0.5cm
\includegraphics[width=8cm]{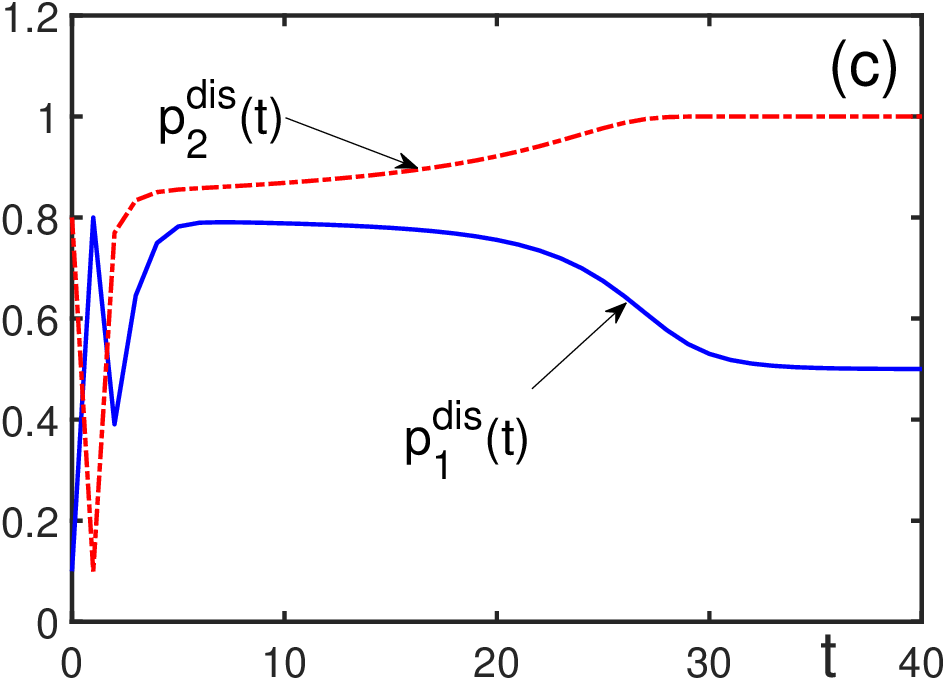} \hspace{5mm}
\includegraphics[width=8cm]{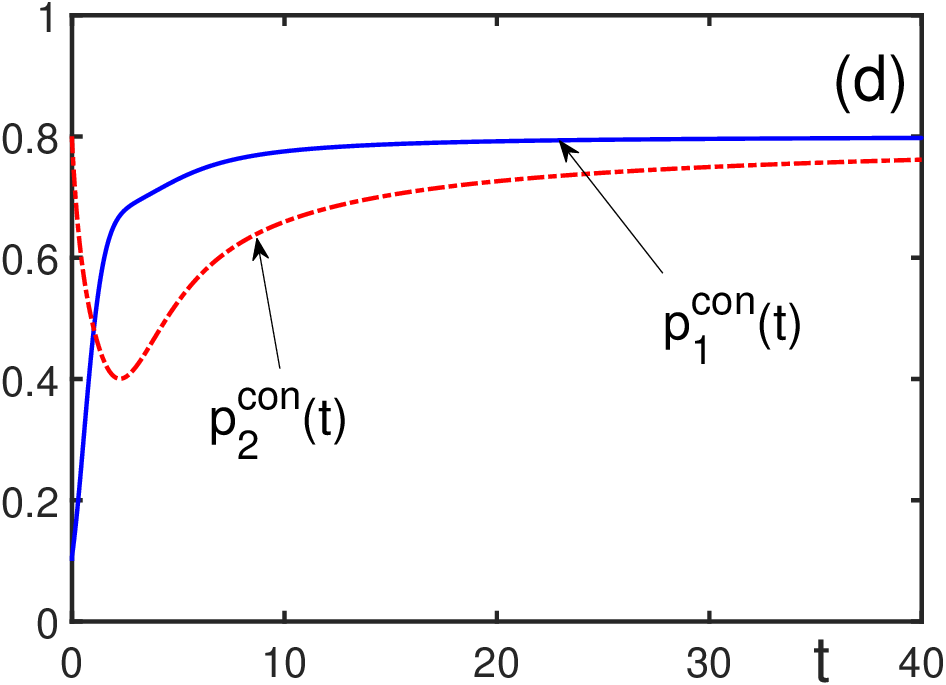}
\caption{\small
Solutions to discrete Equation (\ref{28}) and to continuous Equation 
(\ref{34}) for the initial conditions $f_1=1$, $f_2=0.2$, $q_1=-0.9$, and 
$q_2=0.6$, in the presence of strong herding effect, when $\ep_1=\ep_2=1$:
(\textbf{a}) Discrete solution $p_1^{dis}(t)$ (solid line) tends to the fixed 
point $p_{1dis}^*=0.5$ and continuous solution $p_1^{con}(t)$ (dashed-dotted 
line) tends to the fixed point $p_{1con}^* = f_2 + q_2 = 0.8 = p_{2con}^*$; 
(\textbf{b}) Discrete solution $p_2^{dis}(t)$ (solid line) tends to 
$p_{2dis}^*=1$, while continuous solution $p_2^{con}(t)$ (dashed-dotted line) 
tends to $p_{2con}^* = p_{1con}^* = 0.8$;
(\textbf{c}) Solutions $p_1^{dis}(t)$ and $p_2^{dis}(t)$; 
(\textbf{d}) Solutions $p_1^{con}(t)$ and $p_2^{con}(t)$.
For $t\ra\infty$, solutions $p_1^{con}(t)$ and $p_2^{con}(t)$ tend to the 
same fixed point $p_{1con}^*=p_{2con}^*=f_2+q_2=0.8$, however solution 
$p_1^{dis}(t)$ tends to $p_{1dis}^*=0.5$, whereas solution $p_2^{dis}(t)$ 
tends to $p_{2dis}^*=1$. Discrete and continuous probabilities, though both 
being stable nodes, but tend to different fixed points.
}
\label{fig:Fig.5}
\end{figure}

Figure \ref{fig:Fig.5} explains that discrete and continuous probabilities, though 
both being stable nodes, tend to different fixed points, which do not coincide. 
This happens in the presence of a strong herding effect. 

\begin{figure}[!ht]
\centering
\includegraphics[width=8cm]{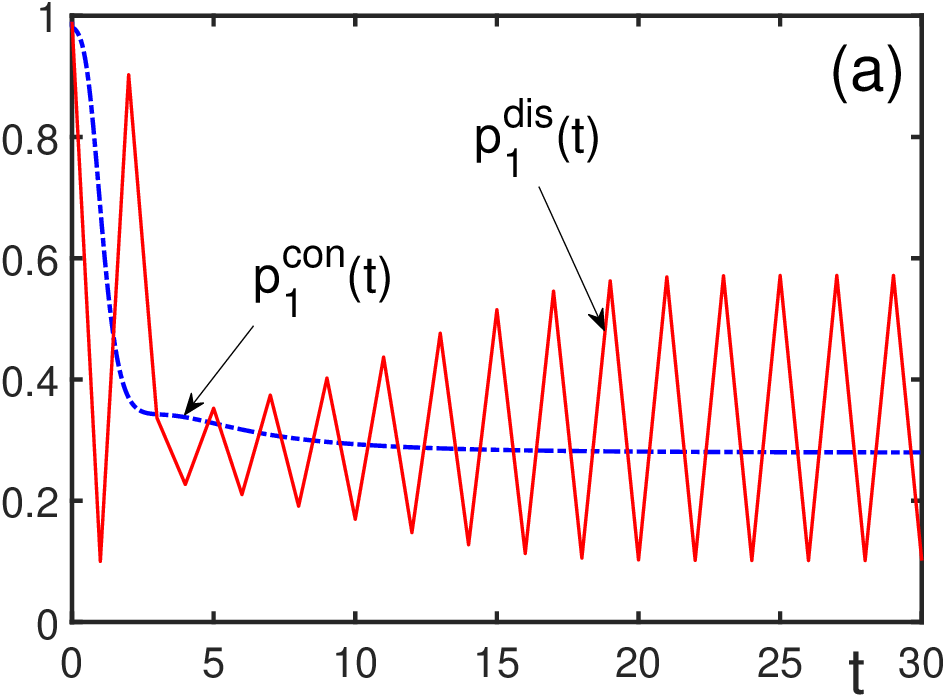} \hspace{5mm}
\includegraphics[width=8cm]{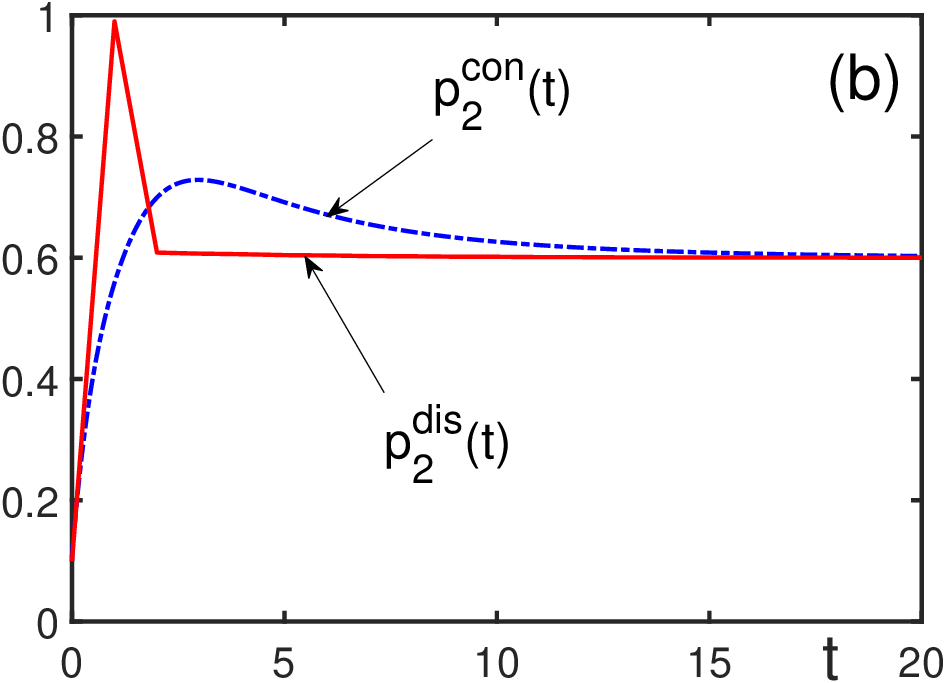}\\
\vskip 0.5cm
\includegraphics[width=8cm]{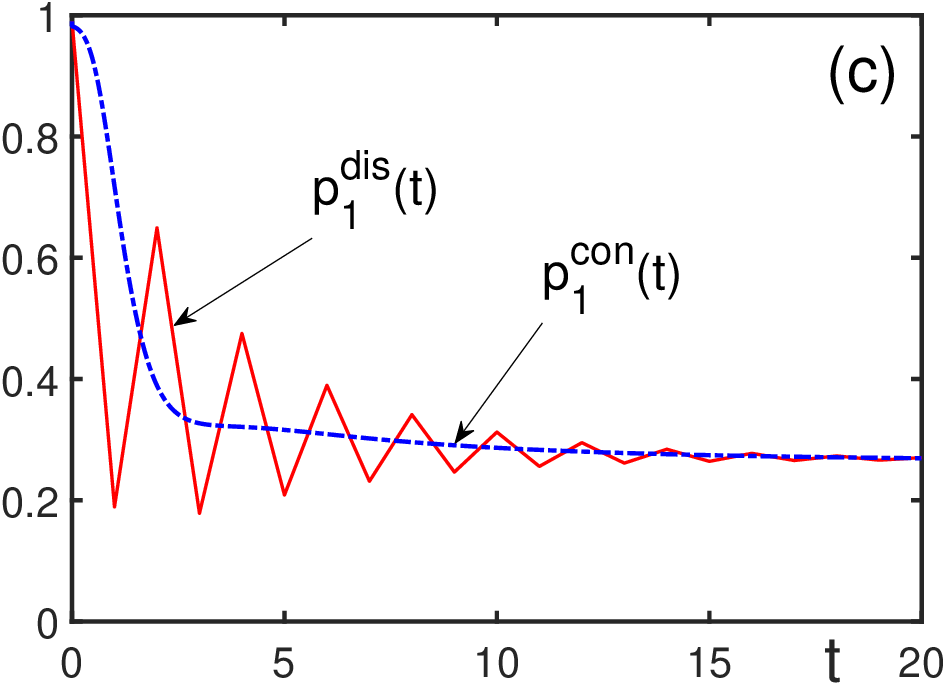} \hspace{5mm}
\includegraphics[width=8cm]{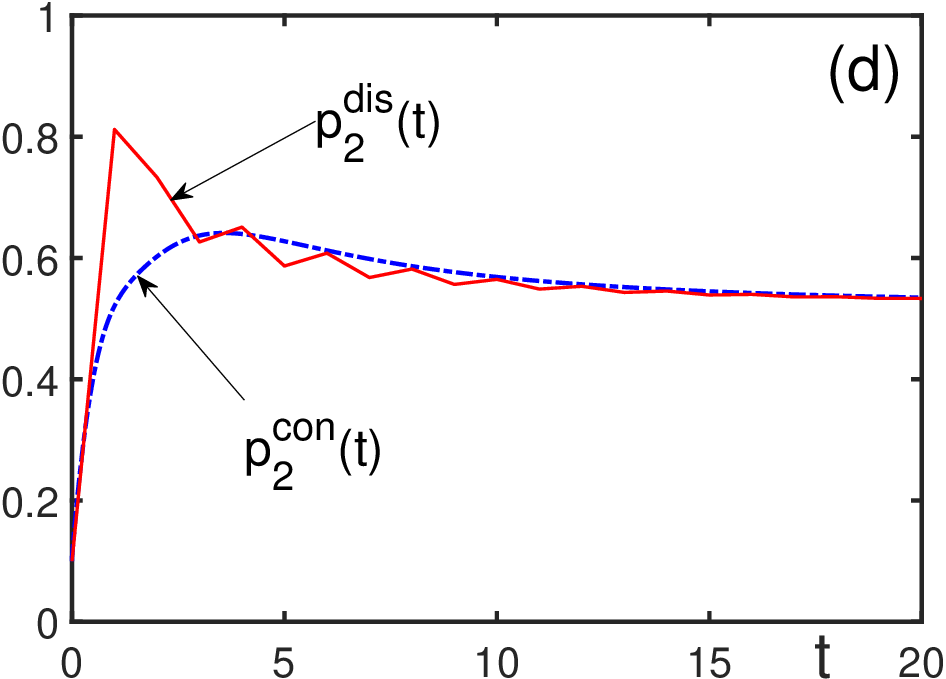}
\caption{\small
Solutions to discrete Equation (\ref{28}) and to continuous Equation 
(\ref{34}) for the initial conditions $f_1=0.6$, $f_2=1$, $q_1=0.39$, and 
$q_2=-0.9$: 
(\textbf{a}) 
Discrete solution $p_1^{dis}(t)$ (solid line) and continuous solution 
$p_1^{con}(t)$ (dashed-dotted line) for the herding parameters $\ep_1=\ep_2=1$. 
Solution $p_1^{con}$(t) tends to the fixed point $p_{1con}^*=0.280$, whereas 
solution $p_1^{dis}(t)$ oscillates with a constant amplitude around $p_{1con}^*$ 
for $t\ra\infty$; 
(\textbf{b}) 
Discrete solution $p_2^{dis}(t)$ (solid line) and continuous solution 
$p_2^{con}(t)$ (dashed-dotted line) for the herding parameters $\ep_1=\ep_2=1$. 
Solutions $p_2^{dis}(t)$ and $p_2^{con}(t)$ tend to the same fixed point 
$p_{2dis}^* =p_{2con}^*=f_1=0.6$; 
(\textbf{c}) 
Discrete solution $p_1^{dis}(t)$ (solid line) and continuous solution 
$p_1^{con}(t)$ (dashed-dotted line) for the herding parameters $\ep_1=0.9$ and 
$\ep_2=0.8$. Solution $p_1^{dis}(t)$ oscillates, and solution $p_1^{con}(t)$ 
monotonically tends to the fixed point $p_{1con}^*=0.265$;
(\textbf{d}) 
Discrete solution $p_2^{dis}(t)$ (solid line) and continuous solution $p_2^{con}(t)$ 
(dashed-dotted line) for the herding parameters $\ep_1=0.9$ and $\ep_2=0.8$. Solution 
$p_2^{dis}(t)$ oscillates, and solution $p_2^{con}(t)$ monotonically tends to the limit 
$p_{2con}^*=0.525$. The behavior of discrete and continuous solutions is qualitatively 
different.
}
\label{fig:Fig.6}
\end{figure}

\begin{figure}[!ht]
\centering
\includegraphics[width=8cm]{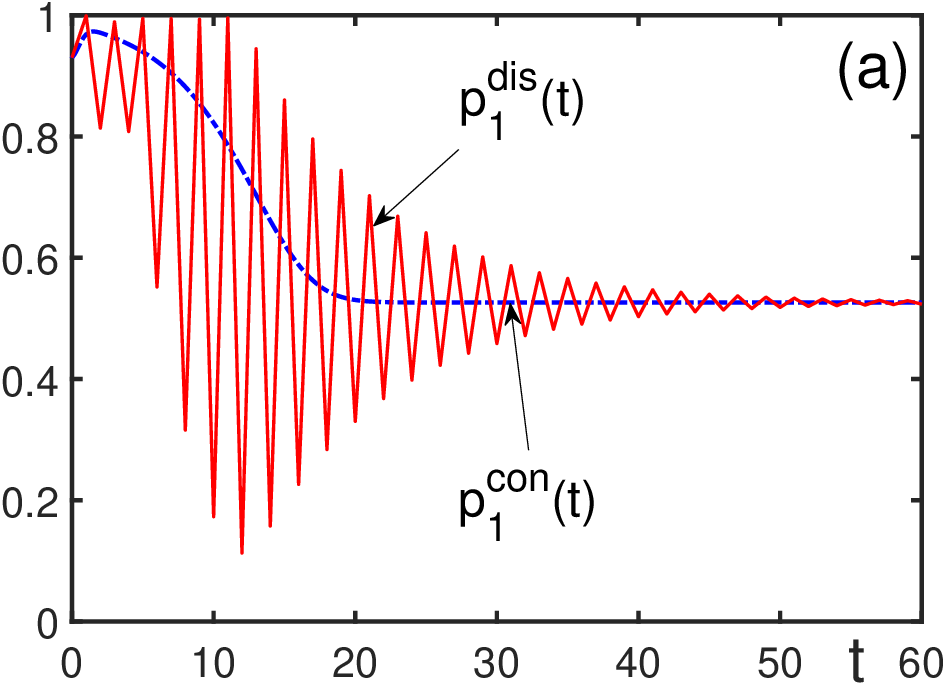} \hspace{5mm}
\includegraphics[width=8cm]{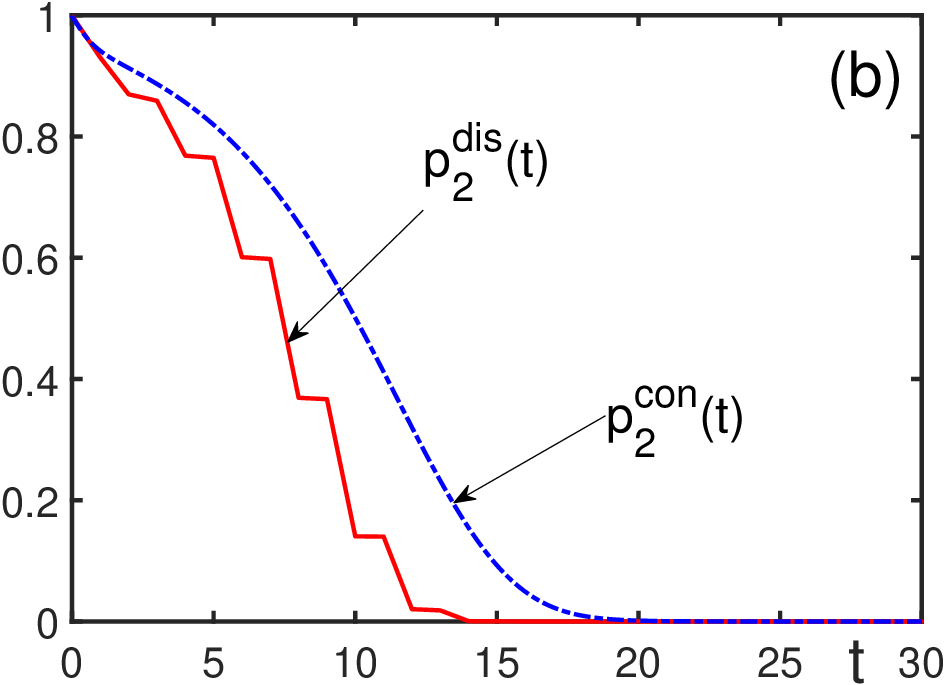}\\
\vskip 0.5cm
\includegraphics[width=8cm]{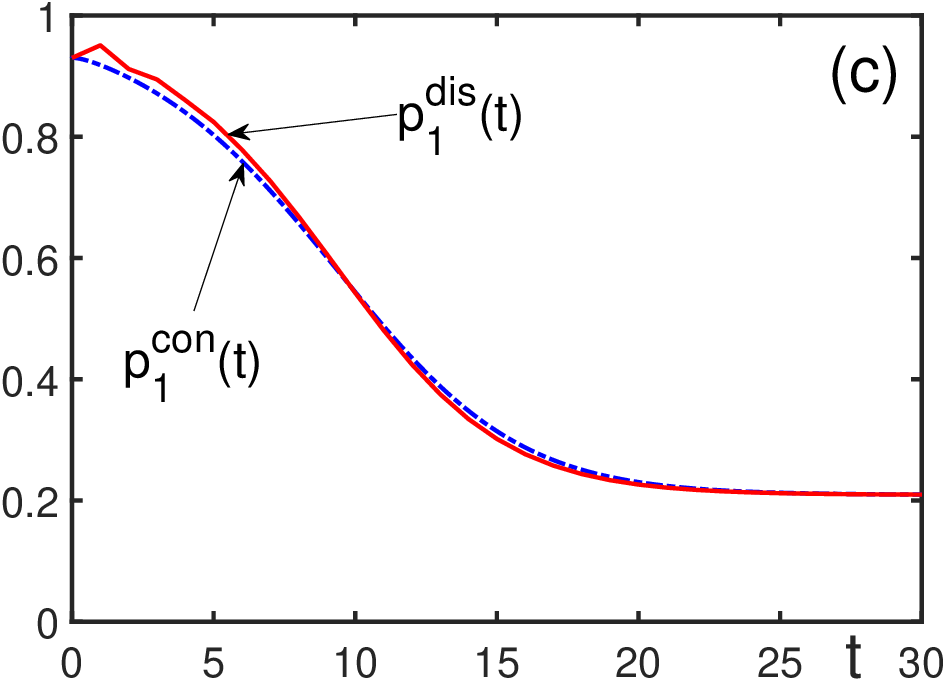} \hspace{5mm}
\includegraphics[width=8cm]{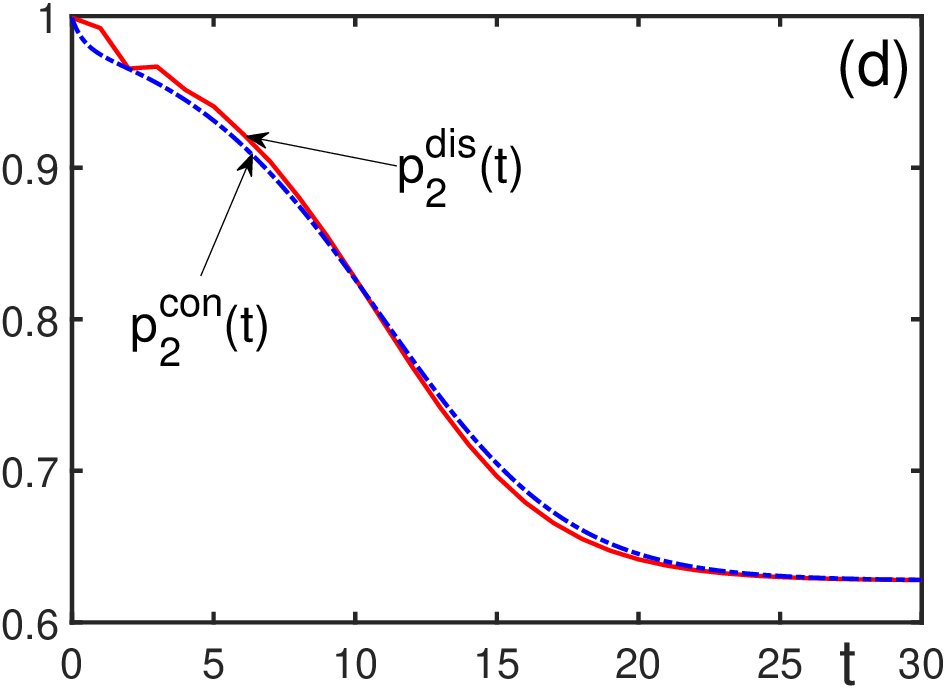}
\caption{\small
Solutions to discrete Equation (\ref{28}) and to continuous Equation (\ref{34}) 
for the initial conditions $f_1=0$, $f_2=0.1$, $q_1=0.93$, and $q_2=0.899$:
(\textbf{a}) 
Discrete solution $p_1^{dis}(t)$ (solid line) and continuous solution 
$p_1^{con}(t)$ (dashed-dotted line) for the herding parameters $\ep_1=\ep_2=1$. 
Solutions $p_1^{con}(t)$, monotonically, and $p_1^{dis}(t)$, with oscillations, 
tend to the same limit $p_{1dis}^*=p_{1con}^*=0.526$; 
(\textbf{b}) 
Solutions $p_2^{dis}(t)$ (solid line) and $p_2^{con}(t)$ (dashed-dotted line) for 
the herding parameters $\ep_1=\ep_2=1$. Solutions $p_2^{dis}(t)$ and $p_2^{con}(t)$ 
tend to the same limit $p_{2dis}^*= p_{2con}^*=f_1=0$;
(\textbf{c}) 
Solutions $p_1^{dis}(t)$ (solid line) and $p_1^{con}(t)$ (dashed-dotted line) 
for the herding parameters $\ep_1=0.3$ and $\ep_2=0.1$. Solution $p_1^{dis}(t)$, 
and solution $p_1^{con}(t)$ monotonically tend to the same limit 
$p_{1dis}^*=p_{1con}^*=0.209$;
(\textbf{d}) 
Solutions $p_2^{dis}(t)$ (solid line) and $p_2^{con}(t)$ (dashed-dotted line) 
for the herding parameters $\ep_1=0.3$ and $\ep_2=0.1$. Solution $p_2^{dis}(t)$, 
and solution $p_2^{con}(t)$ monotonically tend to the same limit 
$p_{2dis}^*=p_{2con}^* =0.628$. Discrete and continuous probabilities tend to 
common fixed points, but in a different way.
}
\label{fig:Fig.7}
\end{figure}

Figures \ref{fig:Fig.6} and \ref{fig:Fig.7} illustrate qualitatively different 
behaviors of discrete and continuous solutions in the presence of the herding 
effect, when the related $p_j^{dis}(t)$ and $p_j^{con}(t)$ can either tend to 
coinciding stable nodes or $p_j^{dis}(t)$ can exhibit oscillations, while 
$p_j^{con}(t)$ smoothly tends to a stable node. 

\begin{figure}[!ht]
\centering\includegraphics[width=8cm]{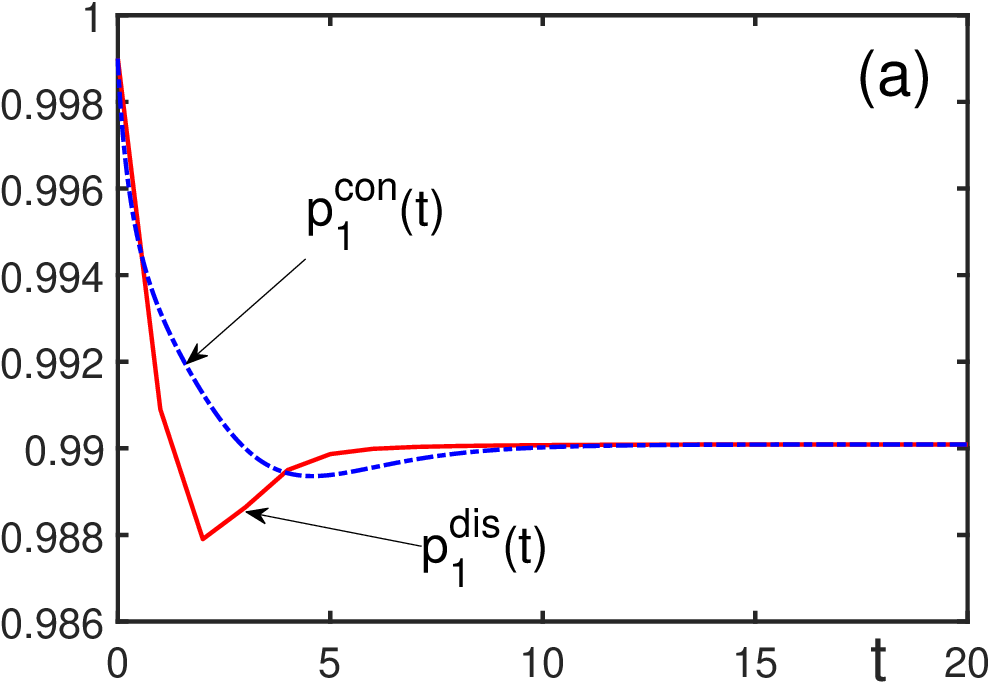} \hspace{5mm}
\includegraphics[width=8cm]{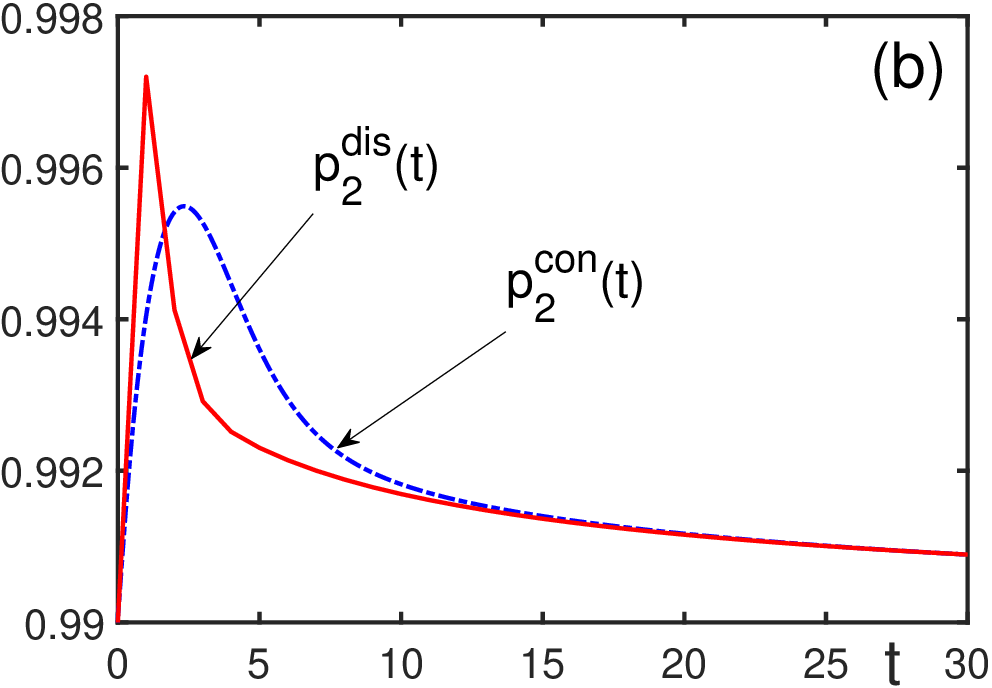}
\caption{\small
Solutions to discrete Equation (\ref{28}) and continuous Equation (\ref{34}) 
for the initial conditions $f_1=0.3$, $f_2=0$, $q_1=0.699$, and $q_2=0.99$, 
with the herding parameters $\ep_1=0.9$ and $\ep_2=0.8$:
(\textbf{a}) 
Solutions $p_1^{dis}(t)$ (solid line) and $p_1^{con}(t)$ (dashed-dotted line). 
Solutions $p_1^{dis}(t)$ and $p_1^{con}(t)$ tend to the same limit $p_1^*=f_2+q_2=0.99$;
(\textbf{b}) 
Solutions $p_2^{dis}(t)$ (solid line) and $p_2^{con}(t)$ (dashed-dotted line). 
Solutions $p_2^{dis}(t)$ and $p_2^{con}(t)$ tend to the same limit $p_2^*=f_2+q_2=0.99$.
Note that here $p_1^*=p_2^*$. All probabilities for the groups with long-term memory
as well as short-term memory, for discrete as well as continuous solutions, tend to the
common fixed point.
}
\label{fig:Fig.8}
\end{figure}

Figure \ref{fig:Fig.8} shows a rare case, where all probabilities for the 
groups with long-term memory as well as short-term memory, for discrete as well 
as continuous solutions, tend to the common fixed point 
$p_{1dis}^* = p_{1con}^* = p_{2dis}^* = p_{2con}^* = f_2 + q_2 = 0.99$.

\begin{figure}[!ht]
\centering\includegraphics[width=8cm]{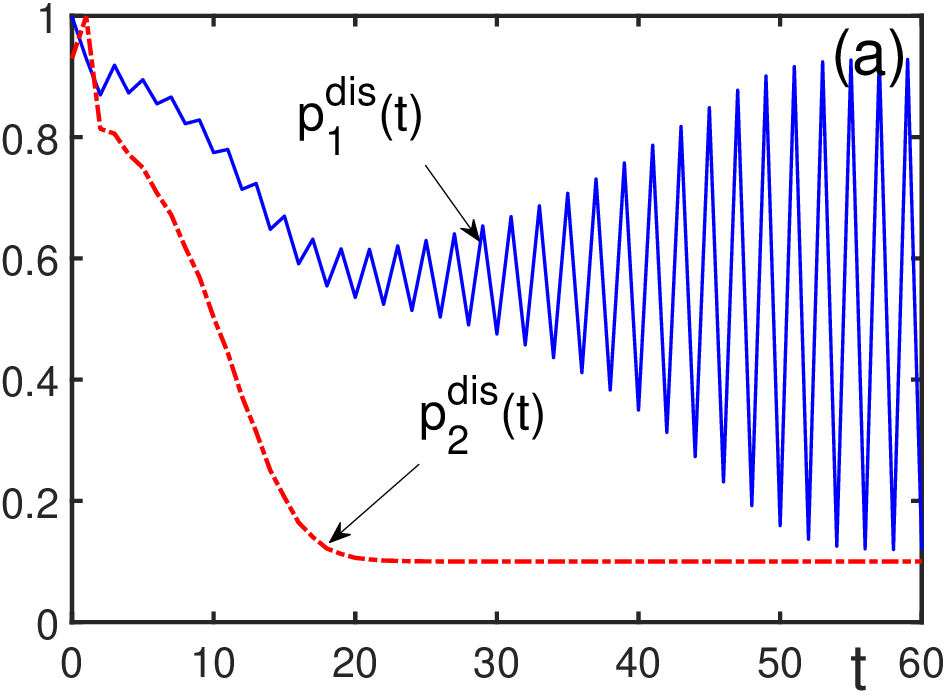} \hspace{5mm}
\includegraphics[width=8cm]{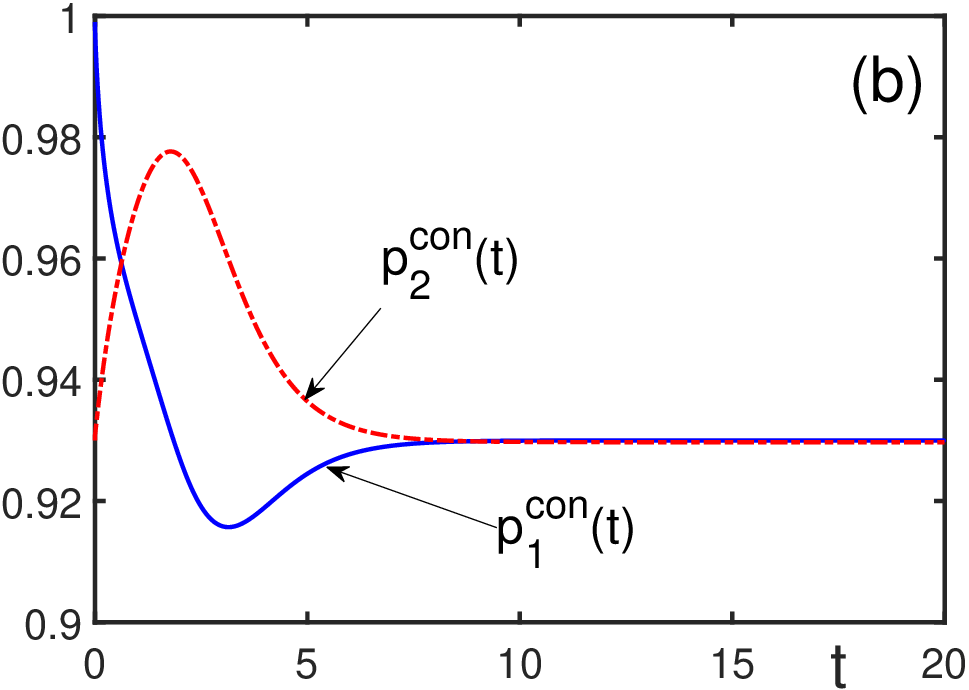}
\caption{\small
Solutions to discrete Equation (\ref{28}) and to continuous Equation (\ref{34}) 
for the initial conditions $f_1=0.1$, $f_2=0$, $q_1=0.899$, and $q_2=0.93$, with 
the herding parameters $\ep_1=\ep_2=1$:
(\textbf{a}) 
Solution to discrete Equation (\ref{28}) $p_1^{dis}(t)$ (solid line) oscillates, 
but solution $p_2^{dis}(t)$ (dashed-dotted line) tends to the fixed point 
$p_2^*=f_1=0.1$; 
(\textbf{b}) 
Solutions to continuous Equation (\ref{34}) $p_1^{con}(t)$ (solid line) and 
$p_2^{con}(t)$ (dashed-dotted line) tend to the same fixed point 
$p_1^*=p_2^*=f_2+q_2=0.93$. Continuous solutions for both groups, with long-term 
and short-term memory, tend to coinciding limits, while the related discrete 
solutions for these groups are very different: One solution permanently oscillates, 
and the other tends to a stable node.
}
\label{fig:Fig.9}
\end{figure}

Figure \ref{fig:Fig.9} gives an example where continuous solutions for both 
groups, with long-term and short-term memory, can tend to coinciding limits, 
while the related discrete solutions for these groups are very different: One 
solution permanently oscillates, and the other tends to a stable node.

\begin{figure}[!ht]
\centering\includegraphics[width=8cm]{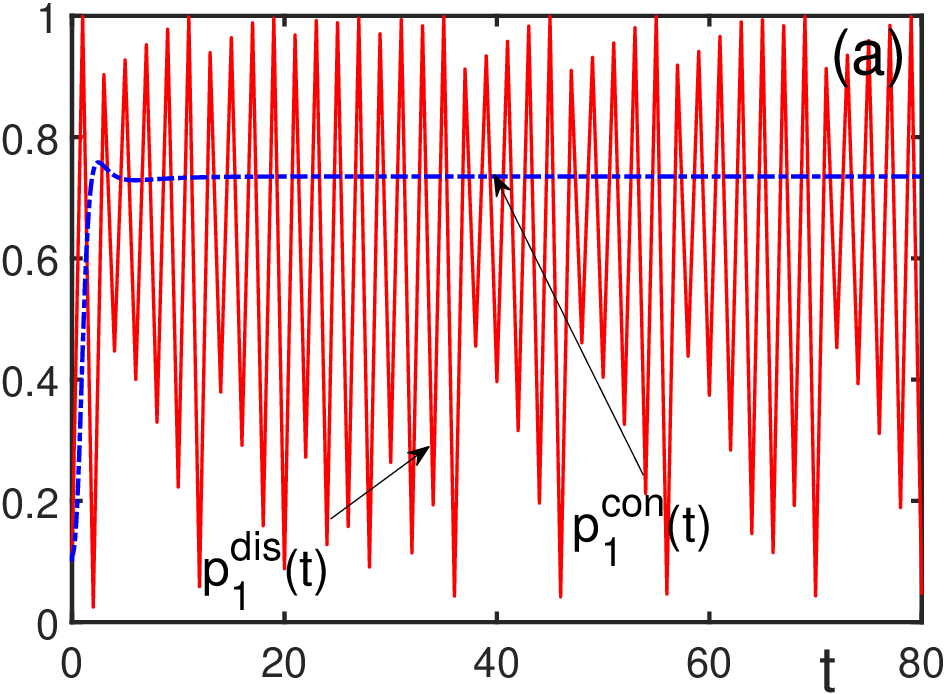} \hspace{5mm}
\includegraphics[width=8cm]{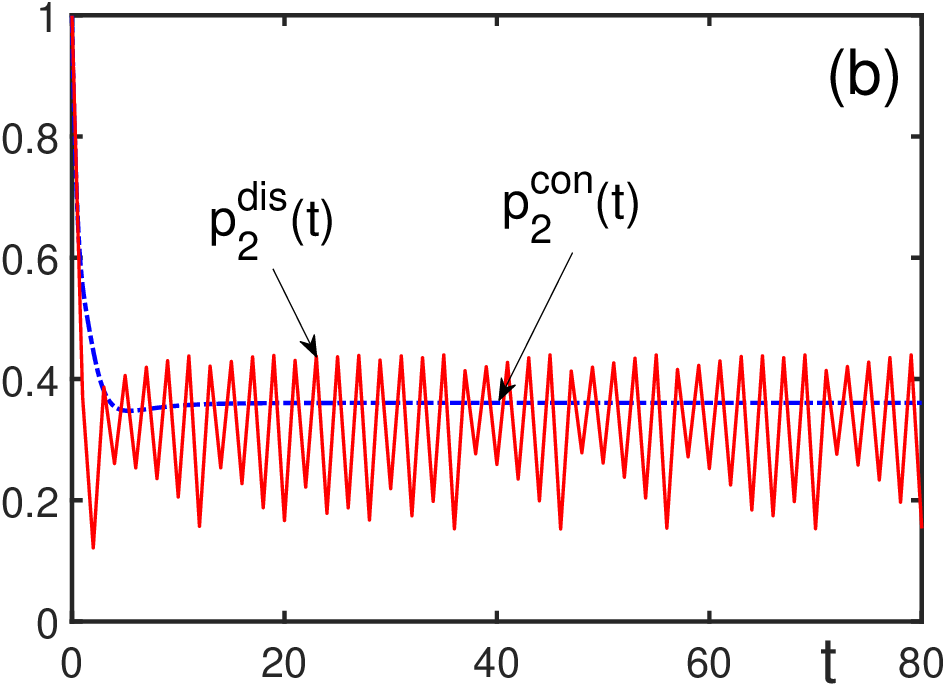}
\caption{\small
Solutions to discrete Equation (\ref{28}) and continuous Equation (\ref{34}) for the
initial conditions $f_1=0.2$, $f_2=0$, $q_1=-0.1$, and $q_2=0.999$, with the herding 
parameters $\ep_1=1$ and $\ep_2=0.7$:
(\textbf{a}) 
Solutions $p_1^{dis}(t)$ (solid line) and $p_1^{con}(t)$ (dashed-dotted line). 
Discrete solution $p_1^{dis}(t)$ chaotically oscillates and continuous solution 
$p_1^{con}(t)$ tends to the limit $p_{1con}^*=0.735$;
(\textbf{b}) 
Discrete solution $p_2^{dis}(t)$ (solid line) and continuous solution $p_2^{con}(t)$ 
(dashed-dotted line). Discrete solution $p_2^{dis}(t)$ chaotically oscillates, while
continuous solution $p_2^{con}(t)$ tends to the limit $p_{2con}^*=0.360$. Discrete 
solutions are chaotic, while, for the same parameters, continuous solutions smoothly 
converge to stable nodes.
}
\label{fig:Fig.10}
\end{figure}

\begin{figure}[!ht]
\centering 
\includegraphics[width=8cm]{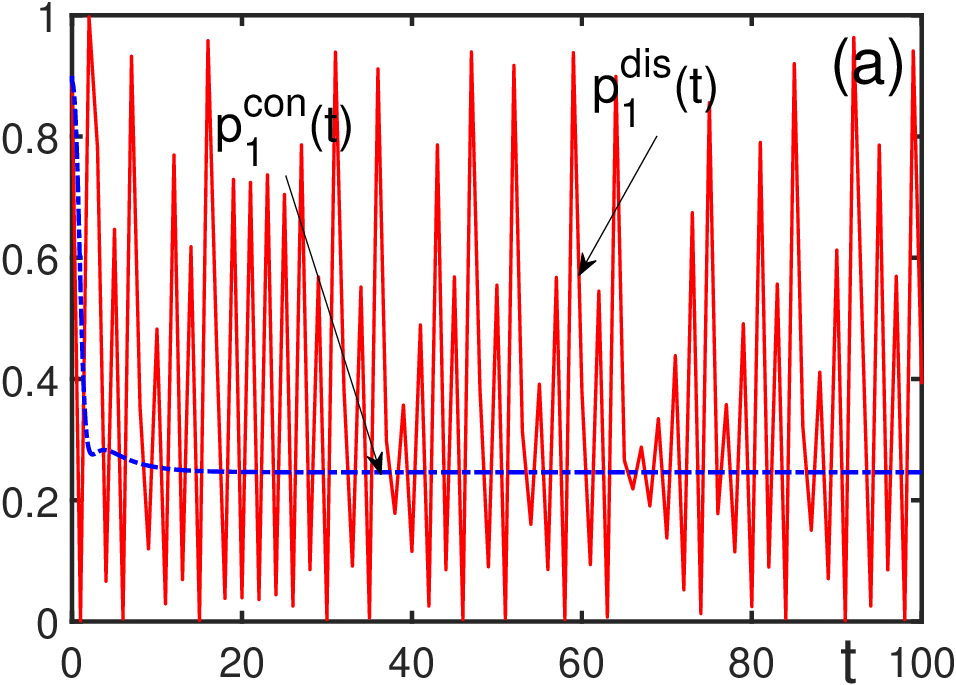} \hspace{5mm}
\includegraphics[width=8cm]{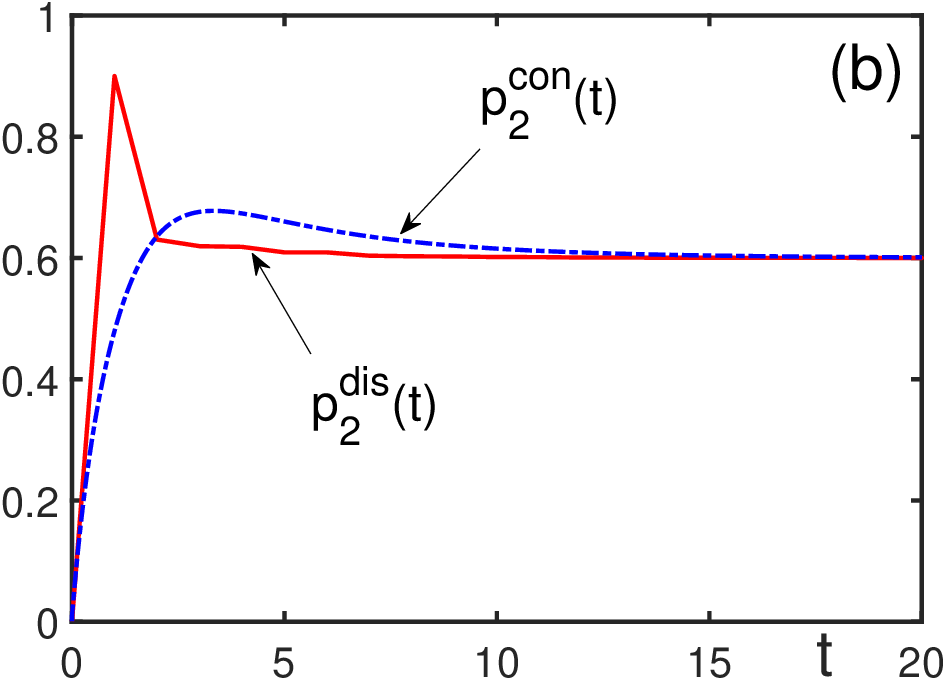}\\
\vskip 0.5cm
\includegraphics[width=8cm]{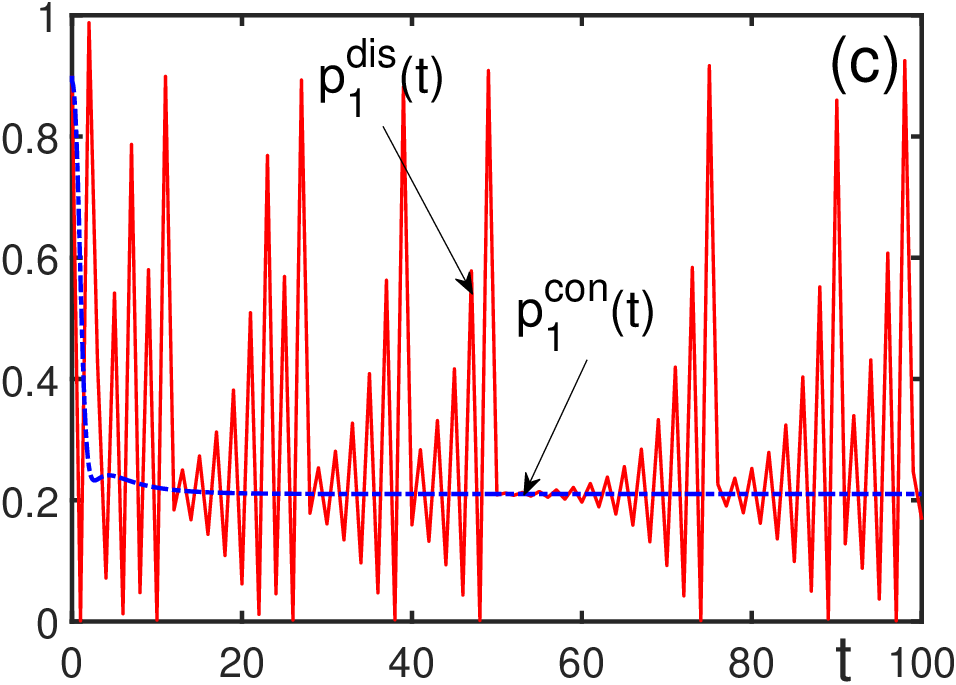} \hspace{5mm}
\includegraphics[width=8cm]{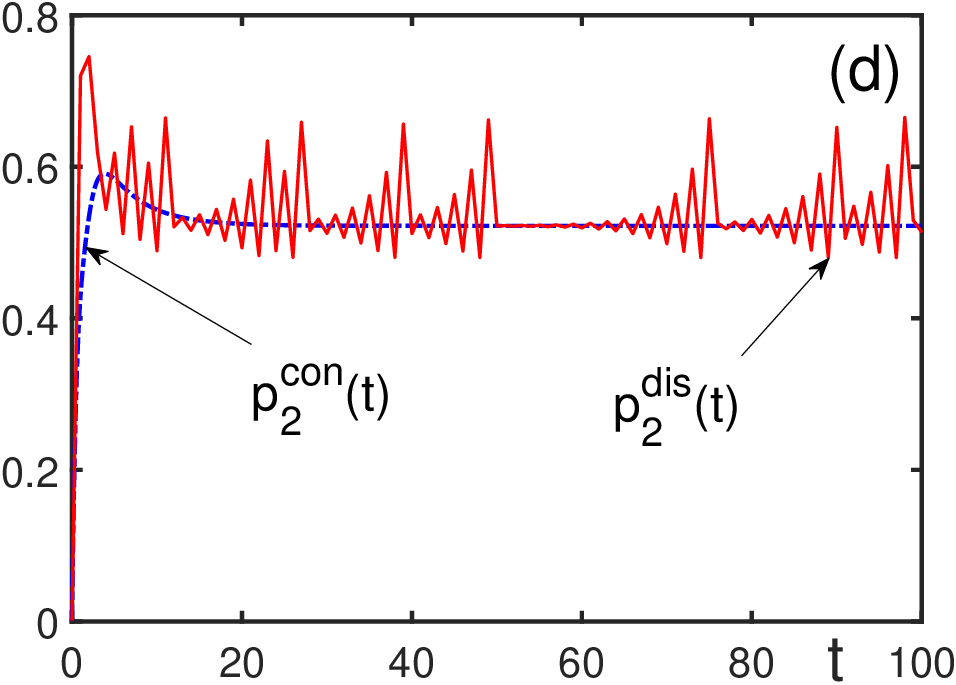}
\caption{\small
Solutions to discrete Equation (\ref{28}) and to continuous Equation (\ref{34}) 
for the initial conditions $f_1=0.6$, $f_2=1$, $q_1=0.3$, and $q_2=-0.999$:
(\textbf{a}) 
Discrete solution $p_1^{dis}(t)$ (solid line) and continuous solution $p_1^{con}(t)$ 
(dashed-dotted line) for the herding parameters $\ep_1=\ep_2=1$. Solution $p_1^{con}(t)$  
tends to the limit $p_{1con}^*=0.246$, while $p_1^{dis}(t)$ chaotically oscillates around 
$p_{1dis}^*$;
(\textbf{b}) 
Discrete solution $p_2^{dis}(t)$ (solid line) and $p_2^{con}(t)$ (dashed-dotted line) 
for the herding parameters $\ep_1=\ep_2=1$. Solutions $p_2^{dis}(t)$ and $p_2^{con}(t)$
tend to the same limit $p_{2dis}^*= p_{2con}^*= f_1=0.6$;
(\textbf{c}) 
Solutions $p_1^{dis}(t)$ (solid line) and $p_1^{con}(t)$ (dashed-dotted line) for 
the herding parameters $\ep_1=1$ and $\ep_2=0.8$. Solution $p_1^{con}(t)$ tends to the 
limit $p_{1con}^*=0.210$, but solution $p_1^{dis}(t)$ chaotically oscillates for all 
times $t\ra\infty$;
(\textbf{d}) 
Solutions $p_2^{dis}(t)$ (solid line) and $p_2^{con}(t)$ (dashed-dotted line) for 
the herding parameters $\ep_1=1$ and $\ep_2=0.8$. Solution $p_2^{con}(t)$ tends to the 
limit $p_{2con}^*=0.522$, while solution $p_2^{dis}(t)$ chaotically oscillates around 
$p_{2dis}^*$. Examples of chaotic behavior of discrete solutions.
}
\label{fig:Fig.11}
\end{figure}

Finally, Figures \ref{fig:Fig.10} and \ref{fig:Fig.11} demonstrate the possibility 
of chaotic behavior for discrete solutions, when, for the same parameters, continuous 
solutions smoothly converge to stable 
nodes.

Summarizing the possible types of behavior, we see that continuous decision making
always displays smooth behavior of probabilities for both groups, with either long-term 
or short-term memory always converging to a stable node. However, discrete decision
making can exhibit, for the same probabilities, a larger variety of behavior types, which
can be smooth, tending to a stable node, or oscillating, hence tending to a stable focus,
or even chaotic.   
 
As far as the temporal behavior of the probabilities of choosing the related alternatives
for discrete and continuous decision making can be essentially different, the natural 
question arises: Which of the algorithms, discrete or continuous, better corresponds 
to the real decision making of social groups? It seems there are activities, such as car 
driving, where decisions can be well approximated by a continuous process. At the same
time, it looks like such processes can be described by a series of decisions occurring 
discretely, although with rather small time intervals between the subsequent steps. It 
may happen that, despite the small time intervals, the discrete and continuous decision 
algorithms lead to different conclusions. From our point of view, the discrete algorithm 
is preferable, since decisions, anyway, are complex, discrete actions composed of several 
subactions: receiving information, processing this information, and making a decision, 
so that always there is a delay time from the start of receiving information to the moment 
of making a decision. The continuous algorithm can provide a reasonable approximation in 
some cases, although sometimes can result in wrong conclusions.       

When a probability $p_j(t)$ converges to a stable node, the corresponding stationary limit 
$p_j^*$ plays the role of the optimal decision taken after multiple steps of decision making,
including the exchange of information with other agents of all groups, taking account 
of agents' emotions, and the tendency of the agents to herding. When a probability oscillates
either periodically or chaotically, this implies that the agents are not able to come to
a decision, but cannot stop hesitating. There exist numerous examples of chaotic behavior
of decision making in medicine, economics, and different types of management
\cite{Baumol_41,Mayer_42,Richards_43,Radzicki_44,Goldberger_45,Cartwright_46,Levy_47,
Barton_48,Krippner_49,Marion_50,Mckenna_51,Mcbride_52}.   

The mathematical reason why the considered continuous solutions for the probabilities cannot 
display chaos is as follows. The probabilities, by definition, are bounded, hence Lagrange 
stable. Then, for a plane motion, the Poincare--Bendixson theorem tells us that if a 
trajectory of a continuous two-dimensional dynamical system is Lagrange stable, then it 
approaches either a stable node or a limit cycle \cite{Yukalov_40}. However, for discrete 
equations, there is no such theorem, and a discrete dynamical system can exhibit chaos.

\section{Conclusions} \label{sec7} 

We have considered affective dynamic decision making, where there are several groups of 
agents choosing between several alternatives. A multistep process of decision making
takes into account the utility of the alternatives, their attractiveness, and the inclination
of the society members to mimic the actions of others (herding effect). Two possible 
algorithms are compared, one algorithm treating multistep decision making as a sequence
of discrete decisions, while the other algorithm studies the overall process as one
continuous action. Dynamic regimes of both algorithms are thoroughly investigated for the 
case of two alternatives and two groups of agents. One group consists of agents with
long-term memory and the other, of agents with short-term memory. 

It is worth stressing that our aim has been not a study of some specific cases, but the 
general understanding of which of the possible algorithms is more appropriate for the 
description in a wide range of parameters corresponding to different situations.

It turns out that the discrete algorithm exhibits much richer behavior that includes the
tendency to a stable node, or to stable focus, or to chaotic behavior. Contrary to this,
the continuous algorithm always results in the convergence to a stable node. In real 
life, as empirical studies show, chaotic decision making can occur in the presence of 
risk and uncertainty. Therefore, it appears that the discrete algorithm is more general, 
while the continuous algorithm can be treated as an approximation that in some cases gives 
a reasonable description, while in many other cases it is not applicable. Anyway, from 
the physiological point of view, multistep decision making better corresponds to a 
sequence of separate decisions than to a single continuous action.

For clarity, above, we kept in mind the frequentist interpretation of probability as a 
fraction of group members. As far as the decision making of a single agent is also a 
probabilistic process, the theory can also be applied to separate agents possessing
different types of memory.

Instead of separate agents, it is possible to consider the nodes of an intelligent network.
For instance, one can keep in mind a neural network, where neurons exchange information 
in order to come to a state represented by a fixed point. The chaotic performance of an 
intelligent network can be interpreted as due to some uncertainty in the process of choice.
For humans, uncertainty can be caused by the complexity of the studied problem or by defects 
in a neural network. Overall, affective intelligence, whether artificial or natural, seems 
to be better described by discrete algorithms than by their continuous approximations. The 
results of this paper can be useful for the creation of affective artificial intelligence.             
 
The probabilistic model of affective decision making, considered in this paper, can be 
extended in several aspects. It is possible to include into consideration more than two 
groups, for instance differing from each other by memory longevity or by the strength
of mutual interactions in the process of exchanging information. It is also possible to
take into account time discounting diminishing the utility factors with time. These 
extensions are planned for future research.


\begin{thebibliography}{999}

\bibitem{Turkle_1}
Turkle, S. 
{\it The Second Self: Computers and the Human Spirit};
Granada: London, UK, 1984.

\bibitem{Brehmer_2}
Brehmer, B.  
Dynamic decision making: Human control of complex systems.
{\it Psychologica} {\bf 1992}, {\it 81}, 211--241.

\bibitem{Beresford_3}
Beresford, B.; Sloper, T.
{\it Understanding the Dynamics of Decision-Making and Choice:
A Scoping Study of Key Psychological Theories to Inform
the Design and Analysis of the Panel Study};
University of York: York/Heslington, UK, 2008.


\bibitem{Evertsz_4}
Evertsz, R.; Thangarajah, J.; Ly, T.
{\it Practical Modelling of Dynamic Decision Making}; 
Springer: Cham, Switzerland, 2019.

\bibitem{Perc_2013}
Perc, M.; Gomez-Gardenes, J.; Szolnoki, A.; Floria, L.M.; Moreno, Y. 
Evolutionary dynamics of group interactions on structured populations: A review. 
{\it J. Roy. Soc. Interface} {\bf 2013}, {\it 10}, 20120997. 

\bibitem{Perc_2017}
Perc, M.; Jordan, J.J.; Rand, D.G.; Wang, Z.; Boccaletti, S.; Szolnoki, A. 
Statistical physics of human cooperation. 
{\it Phys. Rep.} {\bf 2017}, {\it 687}, 1--51. 

\bibitem{Capraro_2021}
Capraro, V.; Perc, M.
Mathematical foundations of moral preferences.
{\it J. R. Soc. Interface} {\bf 2021}, {\it 18}, 20200880.

\bibitem{Jusup_2022}
Jusup, M.; Holme, P.; Kanazawa, K.; Takayasu, M.; Romic, I.; Wang, Z.; Gecek, S.; Lipic, T.; 
Podobnik, B.; Wang, L.; Luo W.; Klanj\u{s}\u{c}ek, T.; Fan, J.; Boccaletti, S.; Perc, M.
Social physics. 
{\it Phys. Rep.} {\bf 2022}, {\it 948}, 1--148.

\bibitem{Yukalov_5}
Yukalov, V.I.
A resolution of St. Petersburg paradox.
{\it J. Math. Econ.} {\bf 2021}, {\it 97}, 102537.

\bibitem{Yukalov_6}
Yukalov, V.I.
Quantification of emotions in decision making.
{\it Soft Comput.} {\bf 2022}, {\it 26}, 2419--2436.

\bibitem{Yukalov_7}
Yukalov, V.I.
Quantum operation of affective artificial intelligence.
{\it Laser Phys.} {\bf 2023}, {\it 33}, 065204.

\bibitem{Gonzalez_8}
Gonzalez, C.; Vanyukov, P.; Martin, M.K. 
The use of microworlds to study dynamic decision making.
{\it Comput. Human Behav.} {\bf 2005}, {\it 21}, 273--286.

\bibitem{Barendregt_9}
Barendregt, N.W.; Josi\'{c}, K.; Kilpatrick, Z.P.
Analyzing dynamic decision-making models using Chapman-Kolmogorov equations.
{\it J. Comput. Neurosci.} {\bf 2019}, {\it 47}, 205–222.

\bibitem{Behrens_10}
Behrens, T.E.; Woolrich, M.W.; Walton, M.E.; Rushworth, M.F.
Learning the value of information in an uncertain world.
{\it Nature Neurosci.} {\bf 2007}, {\it 10}, 1214.

\bibitem{Ossmy_11}
Ossmy, O.; Moran, R.; Pfeffer, T.; Tsetsos, K.; Usher, M.; Donner, T.H. 
The timescale of perceptual evidence integration can be adapted to the 
environment. 
{\it Current Biol.} {\bf 2013}, {\it 23}, 981--986.

\bibitem{Yu_12}
Yu, A.J.; Cohen, J.D. 
Sequential effects: Superstition or rational behavior? 
{\it Adv. Neural Inform. Process.} {\it Systems} {\bf 2008}, {\it 21}, 1873--1880.

\bibitem{Brea_13}
Brea, J.; Urbanczik, R.; Senn, W. 
A normative theory of forgetting: Lessons from the fruit fly. 
{\it PLoS Comput. Biol.} {\bf 2014}, {\it 10}, 1003640.

\bibitem{Urai_14}
Urai, A.E.; Braun, A.; Donner, T.H. 
Pupil-linked arousal is driven by decision uncertainty and alters serial choice bias. 
{\it Nature Commun.} {\bf 2017}, {\it 8}, 14637.

\bibitem{Baddeley_15}
Baddeley, A. 
{\it Working Memory, Thought, and Action}; 
Oxford University Press: Oxford, UK, 2007. 

\bibitem{Albrecht_2023}
Albrecht, S.V.; Christianos, F.; Sch\"{a}fer, L.
{\it Multi-Agent Reinforcement Learning: Foundations and Modern Approaches};
Massachusetts Institute of Technology: Cambridge, MA, USA, 2023.

\bibitem{Neumann_69}
von Neumann, J.; Morgenstern, O. 
{\it Theory of Games and Economic Behavior};
Princeton University Press: Princeton, NJ, USA, 1953.

\bibitem{Savage_70}
Savage, L.J. 
{\it The Foundations of Statistics};
Wiley: New York, NY, USA, 1954.

\bibitem{Kurtz_71}
Kurtz-David, V.; Persitz, D.; Webb, R.; Levy, D.J.
The neural computation of inconsistent choice behaviour.
{\it Nature Commun.} {\bf 2019}, {\it 10}, 1583.

\bibitem{Yaari_72}
Yaari, M.E. 
The dual theory of choice under risk.
{\it Econometrica} {\bf 1987}, {\it 55}, 95--115.

\bibitem{Reynaa_73}
Reynaa, V.F.; Brainer, C.J.
Dual processes in decision making and developmental neuroscience: A fuzzy-trace model.
{\it Developm. Rev.} {\bf 2011}, {\it 31}, 180--206.

\bibitem{Woodford_74}
Woodford, M.
Modeling imprecision in perception, valuation and choice.
{\it Annu. Rev. Econ.} {\bf 2020}, {\it 12}, 579--601.

\bibitem{Luce_75}
Luce, R.D. 
{\it Individual Choice Behavior: A Theoretical Analysis};
Wiley: New York, NY, USA, 1959.

\bibitem{Luce_76}
Luce, R.D.; Raiffa, R.
{\it Games and Decisions: Introduction and Critical Survey};
Dover: New York, NY, USA, 1989.

\bibitem{Brandt_77}
Brandt, R.B.
The concept of rational belief.
{\it Monist} {\bf 1985}, {\it 68}, 3--23.

\bibitem{Swinburne_78}
Swinburne, R.
{\it Faith and Reason};
Oxford University: Oxford, UK, 2005.

\bibitem{Steuer_79}
Steuer, R.E. 
{\it Multiple Criteria Optimization: Theory, Computation and Application};
Wiley: New York, NY, USA, 1986.

\bibitem{Triantaphyllou_80}
Triantaphyllou, E. 
{\it Multi-Criteria Decision Making: A Comparative Study};
Kluwer: Dordrecht, The Netherlands, 2000.

\bibitem{Koksalan_81}
K\"{o}ksalan, M.; Wallenius, J.; Zionts, S. 
{\it Multiple Criteria Decision Making: From Early History to the 21st Century};
World Scientific: Sinapore, 2011.

\bibitem{Basilio_82}
Basilio, M.P.; Pereira, V.; Costa, H.G.; Santos, M.; Ghosh, A.
A systematic review of the applications of multi-criteria decision aid methods (1977--2022).
{\it Electronics} {\bf 2022}, {\it 11}, 1720.

\bibitem{Yukalov_16}
Yukalov, V.I.; Yukalova, E.P.; Sornette, D. 
Information processing by networks of quantum decision makers. 
{\it Physica A} {\bf 2018}, {\it 492}, 747--766. 

\bibitem{Yukalov_17}
Yukalov, V.I.; Yukalova, E.P.; Sornette, D. 
Role of collective information in networks of quantum operating agents. 
{\it Physica A} {\bf 2022}, {\it 598}, 127365. 

\bibitem{Yukalov_18}
Yukalov, V.I.; Yukalova, E.P. 
Self-excited waves in complex social systems. 
{\it Physica D} {\bf 2022}, {\it 433}, 133188.

\bibitem{Martin_19}
Martin, E.D. 
{\it The Behavior of Crowds: A Psychological Study}; 
Harper $\&$ Brothers: New York, NY, USA, 1920. 

\bibitem{Sherif_20}
Sherif, M. 
{\it The Psychology of Social Norms}; 
Harper $\&$ Brothers: New York, NY, USA, 1936. 

\bibitem{Smelser_21}
Smelser, N.J. 
{\it Theory of Collective Behavior}; 
Macmillan: New York, NY, USA, 1965. 

\bibitem{Merton_22}
Merton, R.K. 
{\it Social Theory and Social Structure}; 
Macmillan: New York, NY, USA, 1968.

\bibitem{Turner_23}
Turner, R.H.; Killian, L.M. 
{\it Collective Behavior}; 
Prentice-Hall: Englewood Cliffs, NJ, USA, 1993.

\bibitem{Hatfield_24}
Hatfield, E.; Cacioppo, J.T.; Rapson, R.L. 
{\it Emotional Contagion}; 
Cambridge University Press: New York, NY, USA, 1993. 

\bibitem{Brunnermeier_25}
Brunnermeier, M.K. 
{\it Asset Pricing under Asymmetric Information: Bubbles, Crashes, Technical Analysis, 
and Herding}; 
Oxford University Press: New York, NY, USA, 2001. 

\bibitem{Sornette_26}
Sornette, D.
{\it Why Stock Markets Crash};
Princeton University Press: Princeton, NJ, USA, 2003.

\bibitem{Yukalov_27}
Yukalov, V.I.
Selected topics of social physics: Equilibrium systems.
{\it Physics} {\bf 2023}, {\it 5}, 590--635.

\bibitem{Yukalov_28}
Yukalov, V.I.; Sornette, D. 
Manupulating decision making of typical agents. 
{\it IEEE Trans. Syst. Man Cybern. Syst.} {\bf 2014}, {\it 44}, 1155--1168.

\bibitem{Yukalov_29}
Yukalov, V.I.; Sornette, D. 
Quantitative predictions in quantum decision theory. 
{\it IEEE Trans. Syst. Man Cybern. Syst.} {\bf 2018}, {\it 48}, 366--381.

\bibitem{Read_30}
Read, D.; Loewenstein, G.  
Time and decision: Introduction to the special issue.
{\it J. Behav. Decis. Mak.} {\bf 2000}, {\it 13}, 141--144.

\bibitem{Frederick_31}
Frederick, S.; Loewenstein, G.; O'Donoghue, T.
Time discounting and time preference: A critical review. 
{\it J. Econ. Liter.} {\bf 2002}, {\it 40}, 351--401.

\bibitem{Yukalov_53}
Yukalov, V.I.; Sornette, D.
Role of information in decision making of social agents.
{\it Int. J. Inform. Technol. Decis. Mak.} {\bf 2015}, {\it 14}, 1129--1166.

\bibitem{Kuhberger_54}
K\"{u}hberger, A.; Komunska, D.; Perner, J. 
The disjunction effect: Does it exist for two-step gambles? 
{\it Org. Behav. Human Decis. Process.} {\bf 2001}, {\it 85}, 250--264.

\bibitem{Charness_55}
Charness, G.; Rabin, M. 
Understanding social preferences with simple tests.
{\it Quart. J. Econ.} {\bf 2002}, {\it 117}, 817--869.

\bibitem{Cooper_56}
Cooper, D.; Kagel J.
Are two heads better than one? Team versus individual play in signaling games. 
{\it Am. Econ. Rev.} {\bf 2005}, {\it 95}, 477--509.

\bibitem{Blinger_57}
Blinder, A.; Morgan, J. 
Are two heads better than one? An experimental analysis of group versus individual 
decision-making.
{\it J. Money Credit Bank.} {\bf 2005}, {\it 37}, 789--811.

\bibitem{Sutter_58}
Sutter, M.
Are four heads better than two? An experimental beauty-contest game with teams of 
different size.
{\it Econ. Lett.} {\bf 2005}, {\it 88}, 41--46.

\bibitem{Tsiporkova_59}
Tsiporkova, E.; Boeva, V.
Multi-step ranking of alternatives in a multi-criteria and multi-expert decision
making environment.
{\it Inform. Sci.} {\bf 2006}, {\it 176}, 2673--2697.

\bibitem{Charness_60}
Charness, G.; Karni, E.; Levin, D.
Individual and group decision making under risk: An experimental study of Bayesian 
updating and violations of first-order stochastic dominance.
{\it J. Risk Uncert.} {\bf 2007}, {\it 35}, 129--148.

\bibitem{Charness_61}
Charness, G.; Rigotti, L.; Rustichini, A.
Individual behavior and group membership.
{\it Am. Econ. Rev.} {\bf 2007}, {\it 97}, 1340--1352.

\bibitem{Chen_62}
Chen, Y.; Li, S.
Group identity and social preferences.
{\it Am. Econ. Rev.} {\bf 2009}, {\it 99}, 431--457.

\bibitem{Liu_63}
Liu, H.H.; Colman, A.M. 
Ambiguity aversion in the long run: Repeated decisions under risk and uncertainty.
{\it J. Econ. Psychol.} {\bf 2009}, {\it 30}, 277--284.

\bibitem{Charness_64}
Charness, G.; Karni, E.; Levin, D.
On the conjunction fallacy in probability judgement: New experimental evidence regarding Linda.
{\it Games Econ. Behav.} {\bf 2010}, {\it 68}, 551--556.

\bibitem{Sung_65}
Sung, S.Y.; Choi, J.N.
Effects of team management on creativity and financial performance of organizational teams. 
{\it Org. Behav. Human Decis. Process.} {\bf 2012}, {\it 118}, 4--13.

\bibitem{Schultze_66}
Schultze, T.; Mojzisch, A.; Schulz-Hardt, S.
Why groups perform better than individuals at quantitative judgement tasks.
{\it Org. Behav. Human Decis. Process.} {\bf 2012}, {\it 118}, 24--36.

\bibitem{Xu_67}
Xu, Z.
Approaches to multi-stage multi-attribute group decision making.
{\it Int. J. Inf. Technol. Decis. Mak.} {\bf 2011}, {\it 10}, 121--146.

\bibitem{Tapia_68}
Tapia Garcia, J.M.; Del Moral, M.J.; Martinez, M.A.; Herrera-Viedma, E. 
A consensus model for group decision-making problems with interval fuzzy preference relations.
{\it Int. J. Inf. Technol. Decis. Mak.} {\bf 2012}, {\it 11}, 709--725.

\bibitem{Kullback_32}
Kullback, S.; Leibler, R.A. 
On information and sufficiency. 
{\it Ann. Math. Stat.} {\bf 1951}, {\it 22}, 79--86. 

\bibitem{Kullback_33}
Kullback, S. 
{\it Information Theory and Statistics}; 
Peter Smith: Gloucester, MA, USA, 1978.

\bibitem{James_34}
James, W. 
{\it The Principles of Psychology}; 
Holt: New York, NY, USA, 1890.

\bibitem{Fitts_35}
Fitts, P.M.; Posner, M.I. 
{\it Human Performance};
Brooks/Cole: Boston, MA, USA, 1967.

\bibitem{Cowan_36}
Cowan, N.
What are the differences between long-term, short-term, and working memory.
{\it Prog. Brain Res.} {\bf 2008}, {\it 169}, 323--338.

\bibitem{Camina_37}
Camina, E.; G\"{u}ell, F.
The neuroanatomical, neurophysiological and psychological basis of memory: Current 
models and their origins.
{\it Front. Pharmacol.} {\bf 2017}, {\it 8}, 438.

\bibitem{Gershenfeld_38}
Gershenfeld, N.A. 
{\it The Nature of Mathematical Modeling}; 
Cambridge University Press: Cambridge, UK, 1999. 

\bibitem{Matsumoto_39}
Matsumoto, A.; Szidarovszky, F.
{\it Dynamic Oligopolicies with Time Delays};
Springer: Singapore, 2018.

\bibitem{Yukalov_40}
Yukalov, V.I.
Selected topics of social physics: Nonequilibrium systems.
{\it Physics} {\bf 2023}, {\it 5}, 704--751.

\bibitem{Baumol_41}
Baumol, W.; Benhabib, J. 
Chaos: Significance, mechanism, and economic applications. 
{\it J. Econ. Perspect.} {\bf 1989}, {\it 3}, 77--105.

\bibitem{Mayer_42}
Mayer-Kress, G.; Grossman, S. 
Chaos in the international arms race. 
{\it Nature} {\bf 1989}, {\it 337}, 701--704.

\bibitem{Richards_43}
Richards, D.
Is strategic decision making chaotic?
{\it Behav. Sci.} {\bf 1990}, {\it 35}, 219--232.

\bibitem{Radzicki_44}
Radzicki, M.J.  
Institutional dynamics, deterministic chaos, and self-organizing systems. 
{\it J. Econ. Issues} {\bf 1990}, {\it 24}, 57--102.

\bibitem{Goldberger_45}
Goldberger, A.L.; Rigney, D.R.; West, B.J. 
Chaos and fractals in physiology.
{\it Sci. Am.}, {\bf 1990}, {\it 263}, 43--49.

\bibitem{Cartwright_46}
Cartwright, T.J. 
Planning and chaos theory. 
{\it J. Am. Plann. Assoc.} {\bf 1991}, {\it 57}, 44--56.

\bibitem{Levy_47}
Levy, D. 
Chaos theory and strategy: Theory, application, and managerial implications. 
{\it Strateg. Manag. J.} {\bf 1994}, {\it 15}, 167--178.

\bibitem{Barton_48}
Barton, S. 
Chaos, self-organization, and psychology. 
{\it Am. Psychol.} {\bf 1994}, {\it 49}, 5--14.

\bibitem{Krippner_49}
Krippner, S.  
Humanistic psychology and chaos theory: The third revolution and the third force. 
{\it J. Human. Psychol.} {\bf 1994}, {\it 34}, 48--61.

\bibitem{Marion_50}
Marion, R. 
{\it The Edge of Organisations: Chaos and Complexity Theories of Formal Social Systems}; 
Sage Publications: Thousand Oaks, CA, USA, 1999.

\bibitem{Mckenna_51}
McKenna, R.J.; Martin-Smith, B. 
Decision making as a simplification process: New conceptual perspectives. 
{\it Manag. Decis.} {\bf 2005}, {\it 43}, 821--836.

\bibitem{Mcbride_52}
McBride, N. 
Chaos theory as a model for interpreting information systems in organisations. 
{\it Inform. Syst. J.} {\bf 2005}, {\it 15}, 233--254.

\end{thebibliography}
\end{document}